\pdfoutput=1

\documentclass[11pt]{article}

\usepackage{acl}

\usepackage{times}
\usepackage{latexsym}

\usepackage[T1]{fontenc}

\usepackage[utf8]{inputenc}

\usepackage{microtype}

\usepackage{graphicx}

\usepackage{amsmath}

\usepackage{amsfonts}
\usepackage{multirow}
\usepackage{enumitem}
\usepackage{multicol}

\usepackage{booktabs}
\usepackage{bbm}

\newcommand{\multirowcell}[1]{\begin{tabular}[c]{@{}l@{}}#1\end{tabular}}

\graphicspath{ {./figures/} }

\title{Towards Computationally Feasible Deep Active Learning}

\author{
Akim Tsvigun\textsuperscript{1,3 $\diamondsuit$}, Artem Shelmanov\textsuperscript{1,6 $\diamondsuit$}, Gleb Kuzmin\textsuperscript{1,5}, Leonid Sanochkin\textsuperscript{1}, \\ \bf Daniil Larionov\textsuperscript{2,3,5}, 
\bf Gleb Gusev\textsuperscript{1,2,4}, 
 \bf Manvel Avetisian\textsuperscript{1,4}, and Leonid Zhukov\textsuperscript{1,3} \\
 
\textsuperscript{1}AIRI, 
\textsuperscript{2}MIPT, 
\textsuperscript{3}HSE, 
\textsuperscript{4}Sber AI Lab,
\textsuperscript{5}FRC CSC RAS, \\
\textsuperscript{6}ISP RAS Research Center for Trusted Artificial Intelligence \\

\href{mailto:tsvigun@airi.net}{\{tsvigun, shelmanov, kuzmin, sanochkin, gusev, manvel, zhukov\}@airi.net} \\
\href{mailto:dslarionov@isa.ru}{dslarionov@isa.ru}
}

\begin{document}

\maketitle
\begin{abstract}
Active learning (AL) is a prominent technique for reducing the annotation effort required for training machine learning models. Deep learning offers a solution for several essential obstacles to deploying AL in practice but introduces many others. One of such problems is the excessive computational resources required to train an acquisition model and estimate its uncertainty on instances in the unlabeled pool. We propose two techniques that tackle this issue for text classification and tagging tasks, offering a substantial reduction of AL iteration duration and the computational overhead introduced by deep acquisition models in AL. We also demonstrate that our algorithm that leverages pseudo-labeling and distilled models overcomes one of the essential obstacles revealed previously in the literature. Namely, it was shown that due to differences between an acquisition model used to select instances during AL and a successor model trained on the labeled data, the benefits of AL can diminish. We show that our algorithm, despite using a smaller and faster acquisition model, is capable of training a more expressive successor model with higher performance.\footnote{The code for reproducing the experiments is available at \url{https://github.com/AIRI-Institute/al_nlp_feasible}}
\end{abstract}


\section{Introduction}
\let\thefootnote\relax\footnotetext{$\diamondsuit$ Equal contribution, corresponding authors}

Active learning (AL) \cite{cohn1996active} is an approach for reducing the amount of dataset annotation required for achieving the desired level of machine learning model performance.
This is especially important in domains where obtaining labeled instances is expensive or wide crowdsourcing is unavailable. For example, annotation of clinical and biomedical texts usually requires the help of physicians or biomedical researchers. The time of such highly qualified experts is extremely valuable and should be spent wisely. Straightforward annotation of datasets can be very redundant, wasting the time of annotators on unimportant instances. AL alleviates this problem by asking human experts to label only the most informative instances selected according to the information acquired from a machine learning model. The algorithm for selection of such instances is called a \textit{query strategy}, and a model used to estimate the informativeness of yet unlabeled instances is called an \textit{acquisition model}. 

AL starts from a small \textit{seeding set} of labeled instances, which are used to train an initial acquisition model. A query strategy ranks unlabeled instances in a large pool according to a criterion that measures their informativeness based on the acquisition model output. One of the most widely adopted criteria is the uncertainty of the acquisition model on instances in question \cite{lewis1994sequential}. Eventually, top selected instances are presented to annotators, and this active annotation process iteratively continues.

After labels are collected, we would like to train a model for a final application. In the same vein as \cite{lowell2019practical}, we call it a \textit{successor model}. AL can help reduce the amount of annotation required to achieve a reasonable quality of the successor text processing model by multiple times \cite{settles-craven-2008-analysis,Settles2009}.

Recently, deep learning has given us a tool for solving one of the essential problems of AL. When we start annotating, we have to build an acquisition model almost without insights from the data that could help us to do feature engineering or to introduce inductive bias. Deep learning does not require feature engineering and transfer learning with deep pre-trained models like ELMo \cite{peters2018deep}, BERT \cite{Devlin2019BERTPO}, and followers such as ELECTRA \cite{electra} provide near state-of-the-art performance on a variety of tasks without any modifications to their architectures. However, deep learning introduces another problem related to computational performance. Since AL annotation typically is an interactive process, we have to train acquisition models and perform inference on a huge unlabeled pool of instances very quickly. This imposes constraints on the acquisition model size and entails another issue.

Ideally, the architectures of acquisition and successor models should be the same.
\newcite{lowell2019practical} demonstrate 
that when the acquisition model is different from the successor model, the performance of the latter one can degrade compared to the performance of the model trained on the same amount of annotation obtained without AL. 
The performance drop in the case of acquisition-successor mismatch (ASM) raises the question of whether AL is a practical technique at all since the usage of different models on the annotated dataset is a common practice. 
The problem is complicated by  a contradiction between the fact that the acquisition model is required to be as lightweight as possible to mitigate computational overhead and the successor model should be as expressive as possible because we apparently care about the quality of our final application.

In this work, we propose a simple algorithm based on pseudo-labeling and demonstrate that it is able to alleviate the ASM problem. 
Moreover, we show that it is possible to substitute a resource-intensive acquisition model with a smaller one (e.g., take DistilBERT instead of BERT) but train a more powerful successor model of an arbitrary type (e.g., ELECTRA) without loss of quality. This helps to accelerate the execution of AL iterations and reduce computational overhead. 

We also find that the most time-consuming part of an AL iteration with uncertainty-based query strategies can be the inference on the unlabeled pool of instances, while a set of the most certain instances usually does not change substantially from iteration to iteration. Therefore, the straightforward approach to instance acquisition wastes much time on instances shown to be unimportant in previous iterations. We leverage this finding and propose an algorithm that subsamples instances in the unlabeled pool depending on their uncertainty scores obtained on previous AL iterations. This helps to speed up the AL iterations 
further, especially when the unlabeled pool is large. A series of experiments on text classification and tagging benchmarks widely used in recent works on AL demonstrate the efficiency of the proposed algorithms. 

The contributions of the paper are the following:
\begin{itemize}[itemsep=1mm, parsep=0pt]
    \item We propose a novel algorithm denoted as \textbf{P}seudo-\textbf{L}abeling for \textbf{A}cquisition \textbf{S}uccessor \textbf{M}ismatch (PLASM) that allows the use of computationally cheap models during the acquisition of instances in AL, while it does not introduce constraints on the type of the successor model and effectively alleviates the ASM problem. It helps to reduce the hardware requirements and the duration of AL iterations.
    \item We propose a novel algorithm denoted as \textbf{U}nlabeled \textbf{P}ool \textbf{S}ubsampling (UPS) that helps to reduce the time required for calculating informativeness of instances in AL based on the fact that the set of instances that model is certain about does not change substantially. This helps to further speed up the AL iteration.
\end{itemize}



\section{Related Work}\label{sec:relwork}


Deep learning, to a large extent, has freed data scientists from doing feature engineering, which has been one of the essential obstacles to annotation with AL. This advantage has sparked a series of works on deep active learning (DAL) in natural language processing (NLP). 

\newcite{shen2018} conduct one of the first investigations on DAL in sequence tagging tasks. They propose an efficient way of quantifying the uncertainty of sentences, namely maximal normalized log probability (MNLP), by averaging log probabilities of their tokens. They also address the problem of excessive duration of a neural network training step during an AL iteration by interleaving online learning with training 
from scratch. In our work, we take MNLP as a query strategy for experiments on sequence tagging tasks since it has demonstrated a good trade-off between quality and computational performance.
We consider that online learning can potentially be used as a complement to our algorithms. 
Since the most time-consuming part of an AL iteration can be model inference instead of training, in this work, we also pay attention to the acceleration of the inference step.

Several recent publications investigate deep pre-trained models based on the Transformer architecture \cite{vaswani2017attention}, ELMo \cite{peters2018deep}, and ULMFiT \cite{howard2018universal} in AL on NLP tasks 
\cite{prabhu2019sampling,eindoretal2020active,yuanetal2020cold,shelmanov-etal-2021-active}. We continue this line of works by relying on  pre-trained Transformers since this architecture has been shown promising for AL in NLP due to its good qualitative and computational performance.  

A few works have experimented with Bayesian query strategies for AL. \newcite{shen2018}, \newcite{Siddhant2018DeepBA}, \newcite{eindoretal2020active}, and \newcite{shelmanov-etal-2021-active} leverage Monte Carlo dropout \cite{gal2016dropout} for quantifying uncertainty of models. \newcite{Siddhant2018DeepBA} also apply the Bayes by backprop algorithm \cite{blundell2015weight} for performing variational inference of a Bayesian neural network. 
This approach demonstrates the best improvements upon the 
baseline but introduces large computational overhead both for training and uncertainty estimation of a model, as well as the memory overhead for storing parameters of a Bayesian neural network. The query strategies based on Monte Carlo dropout do not affect the model training procedure and do not change the memory footprint. However, they also suffer from slow uncertainty estimation due to the necessity of making multiple stochastic predictions, 
while their empirical evaluations with Transformers in recent works  \cite{eindoretal2020active,shelmanov-etal-2021-active} do not demonstrate big advantages. 
Therefore, we do not use Bayesian query strategies in our experiments and adhere to the classical uncertainty-based query strategies. 

Recently proposed alternatives to uncertainty-based query strategies leverage reinforcement learning and imitation learning \cite{fang2017learning,liu2018learning,vu2019learning,brantley-etal-2020-active}. This series of works aims at constructing trainable policy-based query strategies. 
However, this requires an excessive amount of computation while the transferability of learned policies across domains and tasks is  underresearched. 

Finally, \newcite{lowell2019practical} question the usefulness of AL techniques in general. They demonstrate that due to the ASM problem, AL can be even detrimental to the performance of the successor. This finding is also revealed for classical machine learning models by \newcite{baldridge-osborne-2004-active}, \newcite{tomanek2011inspecting}, \newcite{hu2016active} and supported by experiments with Transformers in \cite{shelmanov-etal-2021-active}.
Our work directly addresses the question raised by \newcite{lowell2019practical} and suggests a simple solution to the ASM problem. Moreover, we combine it with the method proposed by \newcite{shelmanov-etal-2021-active}, who suggest using distilled models for instance acquisition and their teacher models as successors.



	

\section{Background}\label{sec:methods}

This section describes models and AL query strategies used in this work.

\subsection{Query Strategies}

We conduct experiments with four basic AL query strategies. We note that despite their simplicity, these strategies are usually on par with more elaborated counterparts \cite{eindoretal2020active,shelmanov-etal-2021-active,margatina2021bayesian}.

\textit{\textbf{Random sampling}} is used for both text classification and sequence tagging experiments. Applying this strategy means that we do not use AL at all and just emulate that an annotator labels a randomly sampled piece of a dataset.

\textit{\textbf{Least Confident (LC)}} is used for text classification experiments. This strategy sorts texts in the ascending order of their maximum class probabilities given by a machine learning model. Let $y$ be a predicted class of an instance $x$, then $LC_{cls}$ is: 
\[
\text{LC}_{\text{cls}} = 1 - \max_{y} \mathbb{P}\left( y |x\right).
\]
\textit{\textbf{Maximum Normalized Log-Probability (MNLP)}} is proposed by \newcite{shen2018} to mitigate the drawback of the standard LC when it is applied to sequence tagging tasks. Let $y_i$ be a tag of a token $i$, 
let $x_j$ be 
a token $j$ in an input sequence of length $n$. The MNLP score can be formulated as follows:
\[
\text{MNLP}_{\text{ner}} \! = \!-\!\!\!\max _{y_{1}, \ldots, y_{n}} \! \frac{1}{n} \! \sum_{i=1}^{n} \log \mathbb{P}\left[y_{i} | \{y_j\}\!\setminus \! y_i,\!\left\{\mathbf{x}_{j}\right\}\right].
\]
This modified version of LC works slightly better for sequence tagging tasks \cite{shen2018}, and is adopted in many other works on DAL \cite{Siddhant2018DeepBA,erdmann2019,shelmanov-etal-2021-active}.

\textit{\textbf{Mahalanobis Distance (MD)}} between a test instance and the closest class-conditional Gaussian distribution is suggested by \newcite{NEURIPS2018abdeb6f5} for detection of out-of-distribution instances and adversarial attacks. MD is a strong baseline for uncertainty estimation of NLP model predictions \cite{podolskiy2021revisiting} and is also a backbone for other subsequent techniques \cite{zhou-etal-2021-contrastive}. We use it as an informativeness score in AL since previous work shows that MD captures epistemic uncertainty well \cite{podolskiy2021revisiting}. 
\begin{equation}
    \text{MD}_{\text{cls}} = \min_{c \in C}(h_{i}-\mu_{c})^{T}\Sigma^{-1}(h_{i}-\mu_{c}),
\end{equation}
where $h_{i}$ is a hidden representation of a $i$-th instance, $\mu_{c}$ is a centroid of class $c$, and $\Sigma$ is a covariance matrix for hidden representations of training instances.

\subsection{Models}

We use the standard models based on the Transformer architecture \cite{vaswani2017attention}: BERT, RoBERTa \cite{liu2019roberta}, ELECTRA, and XLNet \cite{NEURIPS2019dc6a7e65}.
For supplementary experiments, we also employ two classical neural models: a CNN-BiLSTM-CRF sequence tagging model \cite{ma-hovy-2016-end} and a CNN-based text classification model \cite{le2018convolutional}.

Besides full-fledged Transformers, we leverage their three smaller distilled versions: DistilBERT \cite{sanh2019distilbert}, DistilRoBERTa, and a custom DistilELECTRA trained by ourselves. The distillation procedure aims at creating a smaller-size model (\textit{student}) while keeping the behavior of the original model (\textit{teacher}) by minimizing the distillation loss over the student predictions and soft target probabilities of the teacher \cite{hinton2015distilling}: 
$L_{distil}=-\sum_{i,c} t_{ic} \cdot \log \left(s_{ic}\right)$, 
where $t_{ic}$ and $s_{ic}$ are probabilities estimated by the teacher and the student correspondingly for each instance $i$ and class $c$. Typically, distillation loss is supplemented with additional techniques that help to align a student with a teacher \cite{sanh2019distilbert}.

Distilled models are usually much more compact than their teachers. For example,
DistilBERT reduces the memory footprint by 40\% compared to the original BERT-base. 
It achieves the 60\% speedup, sacrificing only 3\% of its qualitative performance \cite{sanh2019distilbert}. Since the qualitative performance during acquisition is not essential, we would like to use such lightweight models for instance acquisition to reduce AL iteration duration and the requirements for the computational power of the hardware.

\section{Proposed Methods}

This section outlines two proposed algorithms that help to reduce the computational cost of AL.

\subsection{Pseudo-labeling for Acquisition-Successor Mismatch}

We propose a simple algorithm for constructing a successor model of an arbitrary type using AL: 
\textit{Pseudo-Labeling for Acquisition-Successor Mismatch (PLASM)}. The algorithm is designed for reducing the amount of computation required for instance acquisition during AL with uncertainty-based query strategies. 

PLASM leverages the finding of \newcite{shelmanov-etal-2021-active} that the successor model can be trained on instances labeled during AL without a penalty to the quality if its distilled version was used for instance acquisition. 
However, this idea alone does not resolve the question, how we can train new models of arbitrary type on datasets collected via AL \cite{lowell2019practical}. 

The algorithm consists of the following steps: \vspace{-0.2cm}
\begin{enumerate}[itemsep=1mm, parsep=0pt]
    \item Consider we have a resource-intensive pre-trained teacher model (e.g. BERT). We construct a lightweight distilled version of this model (e.g. DistilBERT) using unlabeled data.
    \item We apply a distilled model to perform acquisition during AL for collecting the gold labels.
    \item The collected labels are used for fine-tuning a resource-intensive teacher model of a higher quality than the distilled acquisition model.
    \item The 
    teacher model is used for pseudo-labeling of the whole unlabeled pool of instances.
    \item The automatically acquired annotations are filtered to reduce noise introduced by mistakes of the pseudo-labeling model. 
    In the main experiments, we use TracIn -- a strong and practical method for mislabelled data identification \cite{pruthi2020estimating}. In the ablation study, we also test a simpler solution: filtering instances with high uncertainty of the pseudo-labeling model predictions.
    The fraction of the filtered out instances in both cases is determined from the evaluation score of the pseudo-labeling model on a held-out subset of the training corpus (100\%-score).
    \item Finally, we train a successor model of an arbitrary type on the dataset that contains automatically labeled instances and instances with gold labels obtained from human experts. 
\end{enumerate}

If the teacher model is expressive enough, it will generate reasonable pseudo labels, which can be filtered and reused by another model of a different type and architecture. 
This additional annotation helps to mitigate the performance drop due to ASM and to keep the benefits of AL even when the successor model is more expressive than the model used for pseudo-labeling.
Meanwhile, PLASM helps to reduce the duration of AL iterations similarly to the approach of \newcite{shelmanov-etal-2021-active}, and it does not introduce any additional computational overhead during the annotation process since training the teacher model and pseudo-labeling are performed after the AL annotation is completed.

\subsection{Unlabeled Pool Subsampling}

If the unlabeled pool of instances is large, which is a common situation, and a deep neural network is used as an acquisition model, the most time-consuming step of the AL cycle is the generation of predictions for unlabeled instances, which is necessary for uncertainty-based query strategies (refer to Table \ref{tab:time_agnews}). We note that uncertainty estimates of the most certain instances in the unlabeled pool do not alter substantially across multiple AL iterations (Table \ref{tab:percents1}). This means that AL wastes much time and resources on these unimportant instances.
We claim that it is possible to recalculate uncertainty scores on the current iteration only for the top instances of the unlabeled pool, which were the most uncertain on previous iterations, while not sacrificing the benefits of AL.



We propose an unlabeled pool subsampling (UPS) algorithm, in which uncertainty estimates only for a fraction of instances are updated. 
On the current iteration, we suggest always selecting a fraction of the most uncertain instances on previous iterations equal to $\gamma \in [0,1]$ and sample a small portion of instances with a probability that depends on their rank in a list sorted by their uncertainty. Formally, this can be written as follows.
Let $u$ be the last recalculated uncertainty score of an instance on one of the previous iterations. We order the instances according to this value: $u_0 \leq u_1 \leq \dots \leq u_i \leq \dots \leq u_M$ and denote a normalized rank of an instance as $r_{i}=\frac{i}{M}$.
Let $T>0$ be a ``temperature'' hyperparameter. 
Then the probability of keeping an instance $i$ for recalculation of uncertainty on the current iteration is:
\[
\mathbb{P}(i)\propto \exp\left(-\frac{max(0, r_i-\gamma)}{T}\right).
\]
Sampling certain instances with a non-negative probability instead of just ignoring  them gives a chance of overcoming a situation when an informative instance is occasionally assigned a high certainty score and is never selected ever since. This method is inspired by subsampling techniques used in gradient boosting algorithms for selecting a training subset for decision trees \cite{NIPS20176449f44a,NEURIPS20195c8cb735}.

On initial AL iterations, an acquisition model is trained on an extremely small amount of data, which leads to unreliable uncertainty estimates. To mitigate this problem, we suggest keeping the standard approach to performing instance acquisition on several first iterations and switching to the optimized process later during AL. We also note that interleaving the optimized selection with the standard approach, in which we recalculate the uncertainty for the whole unlabeled pool of instances, can help to keep the high performance of AL.



\section{Experiments}\label{sec:experiments}

\subsection{Experimental Setup}

We follow the common schema of AL experiments adopted in many previous works \cite{settles-craven-2008-analysis,shen2018,Siddhant2018DeepBA,shelmanov-etal-2021-active}. We emulate the AL annotation cycle starting with a small random sample of the dataset used as a seed for the construction of the initial acquisition model. On each iteration, we pick a fraction of top instances from the unlabeled pool sorted using the query strategy and, instead of demonstrating them to annotators, automatically label them according to the gold standard. These instances are removed from the unlabeled pool and added to the training dataset for the next iterations. On each iteration, we train the successor model on the data acquired so far and evaluate it on the whole available test set. Acquisition and successor models are always trained from scratch.
We run several iterations of emulation to build a chart, which demonstrates the performance of the successor depending on the amount of ``labor'' invested into the annotation process. To report standard deviations of scores, we repeat the whole experiment five times with different random seeds. In most experiments, we use LC or MNLP query strategies for classification and sequence tagging correspondingly. Results with MD are presented only in Figure \ref{fig:asm_mahalanobis} in Appendix \ref{app:add_exp_asm}.

For classification, accuracy is used as the evaluation metric. For sequence tagging, we use the strict span-based F1-score \cite{conll2003}.

\begin{figure*}[t]
    \footnotesize
    
    \centering
    \begin{minipage}[ht]{0.49\linewidth}
    \vspace{-0.3cm}
    \center{\includegraphics[width=1\linewidth]{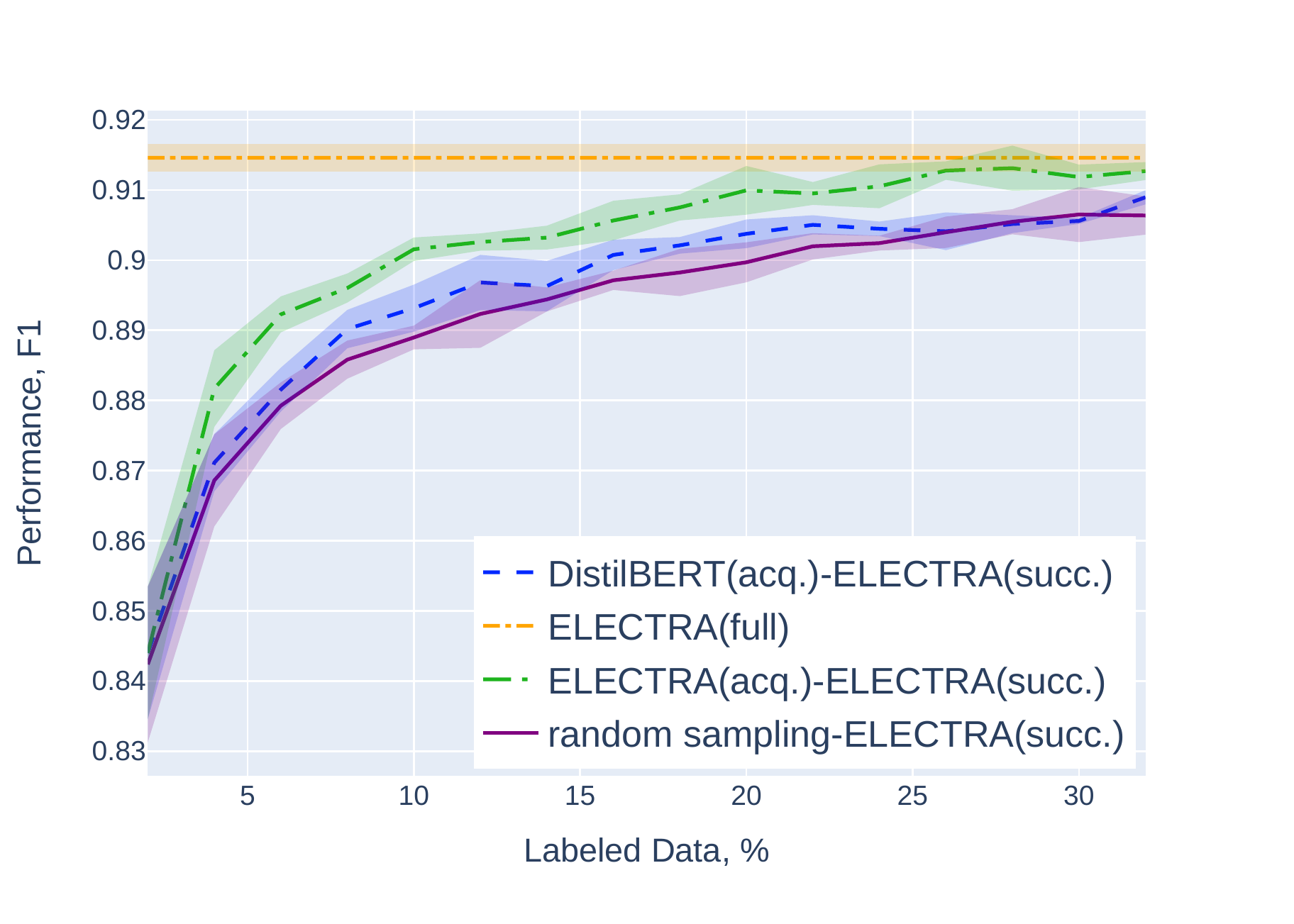} a) ELECTRA is a successor model. }
    \end{minipage}
    \hspace{0.1cm}
    \begin{minipage}[ht]{0.49\linewidth}
    \center{\includegraphics[width=1\linewidth]{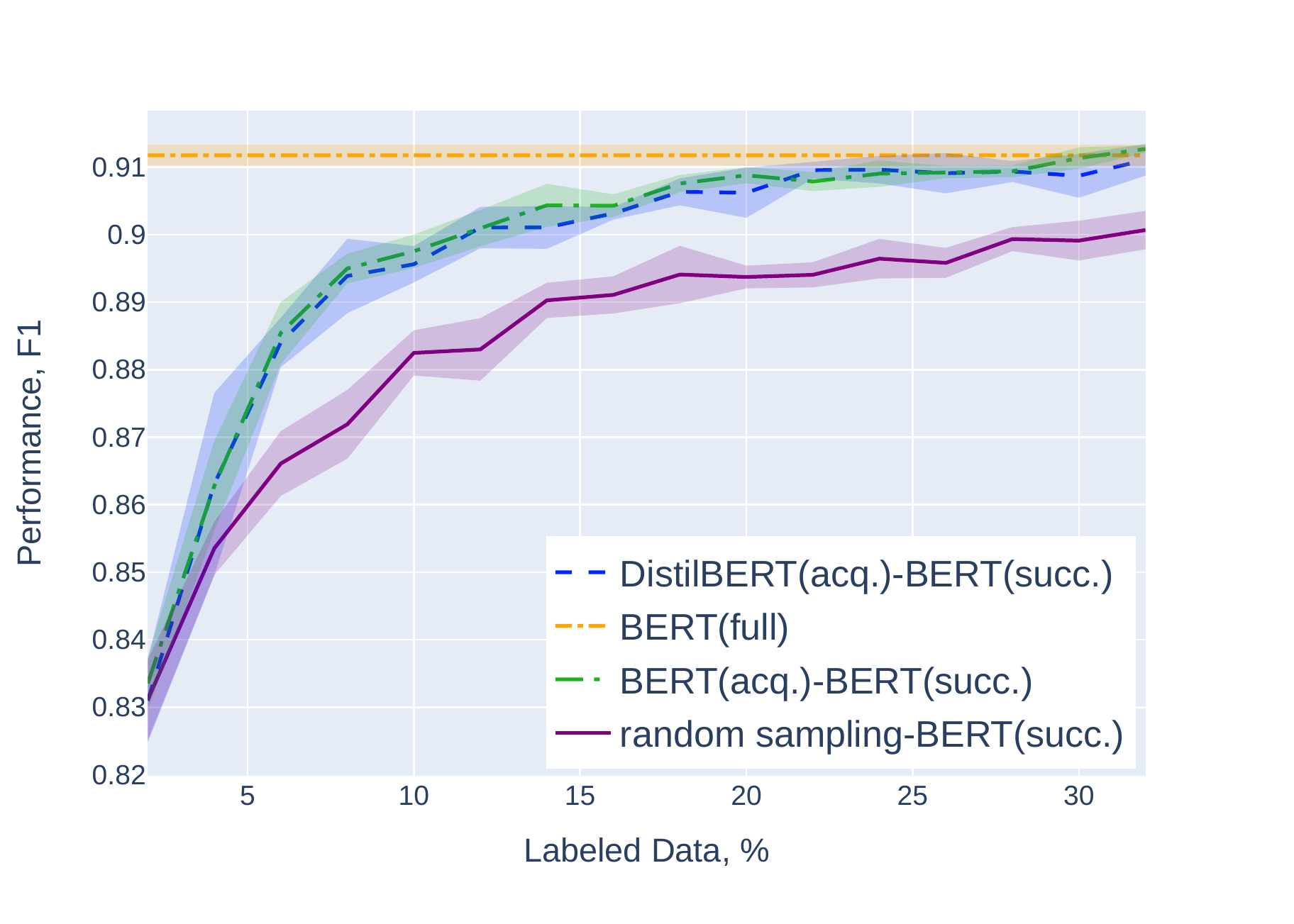} b) BERT is a successor model. \vspace{0.3cm}}
    \end{minipage}
    
    \vspace{-0.2cm}
    \caption{AL experiments on CoNLL-2003, in which a successor model does not match an acquisition model (DistilBERT).}
    \label{fig:asm_conll}
    \vspace{-0.4cm}
    
\end{figure*}

\subsubsection{Datasets}

We experiment with widely-used datasets for the evaluation of AL methods on text classification and sequence tagging tasks. 

For text classification, we use the English AG News topic classification dataset \cite{Zhang2015CharacterlevelCN} and the binary sentiment classification IMDb dataset \cite{maas2011learning}. We randomly select 1\% of instances of the training set as a seed to train the initial acquisition model and select 1\% of instances for ``annotation'' on each AL iteration. 

For sequence tagging, we use English CoNLL-2003 \cite{conll2003} and English OntoNotes 5.0 \cite{pradhan2013towards}. 
We randomly sample instances with a total number of tokens equal to 2\% of all tokens from the training set as a seed. On each AL iteration, we select instances from the unlabeled pool until a total number of tokens equals 2\% of all training tokens. 

The corpora statistics are presented in Table~\ref{tab:corpora1} in Appendix~\ref{app:hyp_data}.


\subsubsection{Model Choice, Training Details, and Hyperparameter Selection}

We conduct experiments with pre-trained Transformers used in several previous works on AL. The exact checkpoints and parameter numbers are presented in Table \ref{tab:checkpoints} in Appendix \ref{app:hyp_data}. Section \ref{app:distillation} contains the distillation details of the custom DistilELECTRA model.



We keep a single pre-selected set of hyperparameters for all AL iterations.
Tables~\ref{tab:hyp}, \ref{tab:bilstm_hp}, \ref{tab:cnn_hp} in Appendix~\ref{app:hyp_data} describe the hyperparameter setup. Hyperparameter tuning on each AL iteration is very time-consuming.
This is an important research problem but out of the scope of the current work.



\begin{figure}[t]
    \vspace{-0.4cm}
	\center{\includegraphics[width=1.\linewidth]{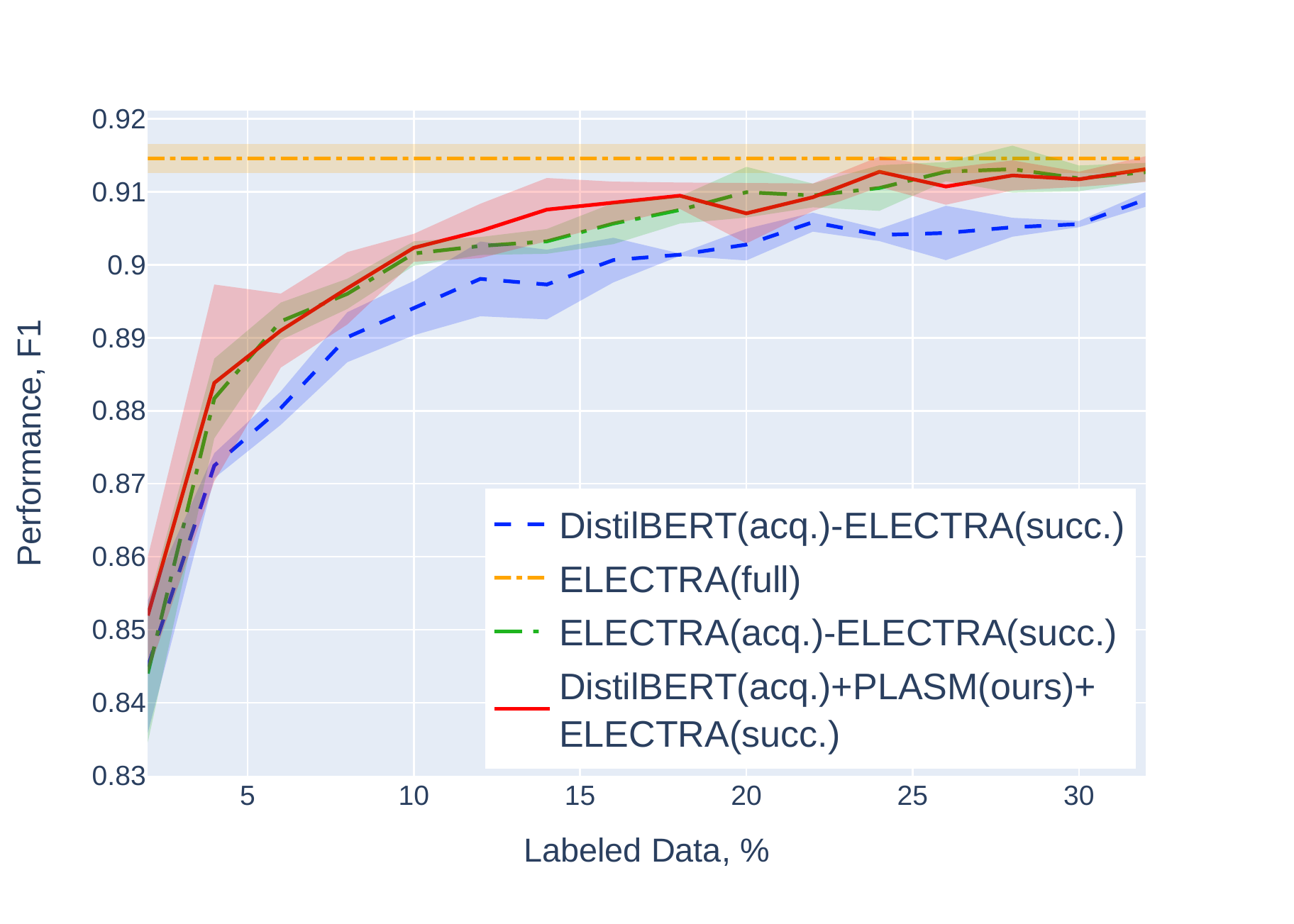}}
	\vspace{-0.6cm}
	\caption{The performance of PLASM (BERT is a pseudo-labeling model) on CoNLL-2003 compared with the standard approach to AL.}
	\label{fig:plasm_conll}
	\vspace{-0.5cm}
\end{figure}

\begin{figure}[t]
	\center{\includegraphics[width=1.\linewidth]{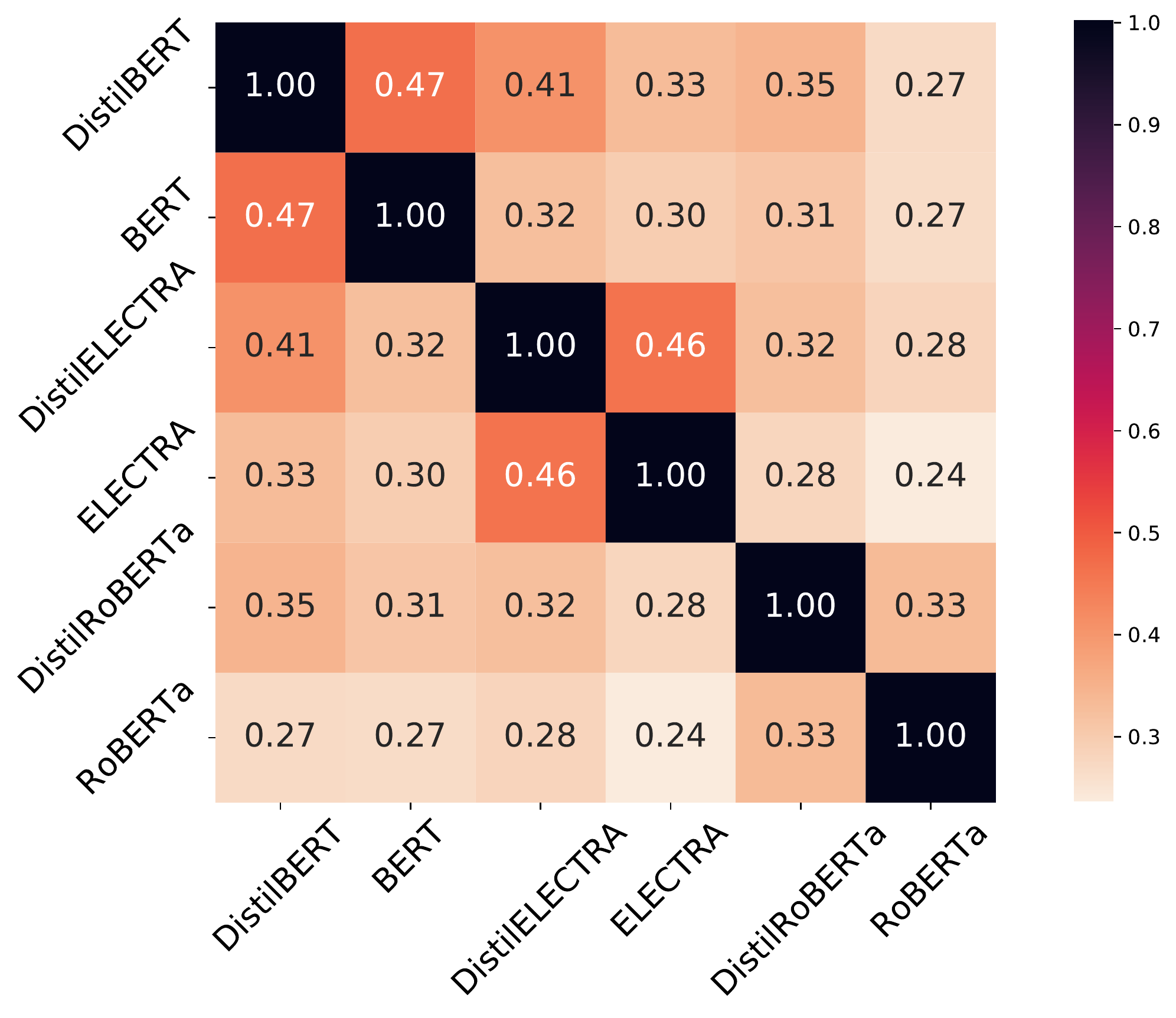}}
	\vspace{-0.6cm}
	\caption{Uncertainty correlation matrix of various Transformers on the AG News dataset. The correlations were obtained by training a model on the 1\% of the training data and calculating the LC score for the rest 99\% of instances.}
	\label{fig:correlation_ue}
	\vspace{-0.5cm}
\end{figure}

\subsection{Results and Discussion}




\subsubsection{Acquisition-Successor Mismatch}

First of all, we illustrate the ASM problem on the selected datasets with various acquisition-successor pairs (Figure \ref{fig:asm_conll}a and Figures \ref{fig:asm_onto}a, \ref{fig:asm_ag}a, \ref{fig:asm_ag_distilelectra}a, \ref{fig:asm_ag_distilroberta}a, \ref{fig:asm_imdb}a in Appendix \ref{app:add_exp_asm}). 
The presented results correspond to the findings of \newcite{lowell2019practical} and \newcite{shelmanov-etal-2021-active}. In each experiment, we see a significant reduction in the performance of successor models when for acquisition, a distilled model from a different family is used.
The performance drop is especially notable when we compare results of ELECTRA(acq.)-ELECTRA(succ.) 
to results of DistilBERT(acq.)-ELECTRA(succ.) on the CoNLL-2003 dataset in Figure \ref{fig:asm_conll}a and to results of DistilRoBERTa(acq.)-ELECTRA(succ.) on AG News in Figure \ref{fig:asm_ag_distilroberta}a in Appendix \ref{app:add_exp_asm}. Moreover, Figure \ref{fig:conll_asm_bert_electra} in Appendix \ref{app:add_exp_asm} shows that even if use 
full-fledged BERT for acquisition and ELECTRA as a successor (and vice versa), a similar performance drop is also present. 

The ASM problem appears to be even more severe with the modern uncertainty estimation technique based on MD. Figure \ref{fig:asm_mahalanobis} in Appendix \ref{app:add_exp_asm} shows the results of experiments with MD and DistilBERT(acq.)-ELECTRA(succ.) on the AG News dataset. On most iterations, the performance drop for MD is even bigger than the drop for LC shown in Figure \ref{fig:asm_ag}a. 

In the next series of experiments, we demonstrate on both text classification and tagging tasks that when an acquisition model is a distilled version of the full-fledged successor model, the ASM problem is substantially alleviated
(Figure \ref{fig:asm_conll}b, and Figures \ref{fig:asm_onto}b, \ref{fig:asm_ag}b, \ref{fig:asm_ag_distilelectra}b, \ref{fig:asm_ag_distilroberta}b, \ref{fig:asm_imdb}b in Appendix \ref{app:add_exp_asm}). Previously, this effect was also revealed by \newcite{shelmanov-etal-2021-active} for tagging. As we can see in Figure \ref{fig:asm_conll}b, when DistilBERT is used as an acquisition model, the successor model based on BERT does not experience a performance drop. A similar effect with other model pairs can be noted for sequence tagging on OntoNotes and for text classification on AG News and IMDb. 

Figure~\ref{fig:correlation_ue} shows the correlation between the output probabilities of various Transformer-based models fine-tuned on 1\% of the AG News dataset. As we can see, for each model, the other most similar model is its student / teacher (except for DistilRoBERTa, which has a slightly larger correlation with DistilBERT). This explains the absence of the ASM problem for such model pairs. Since, after fine-tuning, the distilled version of a model produces similar uncertainty estimates, it will strive to query similar instances during AL. 

Although we can mitigate the ASM problem for such model pairs as DistilBERT-BERT, it is still a serious constraint for applying AL. Obviously, such an approach is not feasible if for the final application, one would like to train a completely different model (e.g. XLNet). In the next section, we show that the proposed method based on pseudo-labeling helps to overcome this limitation and resolve the ASM problem in a more general case. 




\subsubsection{Pseudo-labeling for Acquisition-Successor Mismatch}

Figure~\ref{fig:plasm_conll}, and Figures~\ref{fig:plasm_ag}, \ref{fig:plasm_onto}, \ref{fig:plasm_imdb} in Appendix~\ref{sec:add_exp_plasm} present the performance of PLASM on considered datasets for various combinations of acquisition, successor, and pseudo-labeling Transformer models in comparison with the case when acquisition and successor models are the same and with the case of ASM. On the AG News dataset (Figure \ref{fig:plasm_ag}), we investigate the effect of PLASM for three different successors: ELECTRA, RoBERTa, XLNet and for three different distilled acquisition models: DistilBERT, DistilRoBERTa, and our custom DistilELECTRA model. In all experiments, PLASM substantially alleviates the ASM problem yielding higher results compared to directly fine-tuning on data acquired with a different acquisition model. 

Usually, PLASM yields a similar or slightly better results than the case when the same model is used both for acquisition and as a successor. PLASM might be superior than this case when a pseudo-labeling model is better suited to the dataset than a successor model. For example, in Figure \ref{fig:plasm_ag}d, PLASM shows better results on early AL iterations than ELECTRA(acq.)-ELECTRA(succ.) due to the fact that RoBERTa used for pseudo-labeling has generally higher performance on AG News than ELECTRA when fine-tuned on the same amount of labeled data. However, we argue that PLASM also effectively helps to deal with the ASM problem when the successor model is more expressive than the pseudo-labeling model. This is the case of the experiment on CoNLL-2003 (Figure \ref{fig:plasm_conll}), where PLASM completely mitigates the ASM problem, while ELECTRA successor shows generally better results than BERT used for pseudo-labeling.


Figures~\ref{fig:conll_bilstm} and \ref{fig:agnews_cnn} in Appendix~\ref{sec:add_exp_plasm} show that PLASM also mitigates the performance drop due to ASM between a DistilBERT acquisition model and classical CNN-BiLSTM-CRF or CNN successor models. In this case, PLASM gives a very big boost to performance compared to the case when the same classical model is used both for acquisition and as a successor. This happens because the BERT-based pseudo-labeling model is better suited for fine-tuning on small data than the classical models and produces good automatically labeled instances that are reused by successors. 

Figure~\ref{fig:plasm_ablation}a presents the results of the first ablation study, in which, for pseudo-labeling, we leverage the same distilled model used for acquisition instead of a more expressive teacher. In particular, DistilBERT performs acquisition and pseudo-labeling instead of BERT, while ELECTRA is a successor. The performance drop in this case compared to PLASM demonstrates that using an expressive model (e.g. BERT) for pseudo-labeling is necessary for achieving high scores  
at the beginning of annotation. Figure~\ref{fig:plasm_ablation}b presents the results of the second ablation study, in which we use DistilBERT for acquisition and ELECTRA for pseudo-labeling and as a successor. This study demonstrates that pseudo-labeling on its own cannot alleviate the ASM completely. It is better to use an expressive pseudo-labeling model that also matches the lightweight acquisition model (e.g. distilled model for acquisition, its teacher -- for labeling), as it is proposed in PLASM. 


The ablation study of the methods for filtering erroneous instances in the pseudo-labeling step is conducted on the AG News dataset in Figures \ref{fig:plasm_ablation_filtering}a,b in Appendix \ref{sec:add_exp_plasm}. Applying each of the methods gives substantial improvements over the PLASM without the filtering step, while TracIn is slightly better than thresholding uncertainty of pseudo-labeling model predictions. We note that for XLNet as a successor, PLASM without filtering alleviates the ASM problem, but does not approach the performance of the case when XLNet is used as an acquisition model.

Table~\ref{tab:time_agnews} and Table~\ref{tab:time_conll} in Appendix~\ref{sec:add_exp_ups} summarize the time required for conducting AL iterations with different acquisition functions on the AG News and CoNLL-2003 datasets.
As we can see, since PLASM uses DistilBERT for acquisition, our method reduces the iteration time by more than 30\% compared to the standard approach, in which ELECTRA is used for acquisition. 
Thereby, empirical results show that PLASM offers two benefits: (1) it helps to alleviate the ASM problem in AL; (2)~it reduces the time of an AL iteration and required computational resources for training and running acquisition models. These benefits substantially increase the practicality of using AL in interactive annotation tools.

\begin{table}
    \centering
    \footnotesize
    \scalebox{1.08}{
    \begin{tabular}{cccc}
        \hline
        {\textbf{Top-k\% / Curr. AL iter.}} &      \textbf{1} &      \textbf{2} &      \textbf{6} \\
        \hline
        10\% &  0.503 &  0.649 &  0.924 \\
        20\% &  0.789 &  0.883 &  0.992 \\
        30\% &  0.915 &  0.947 &  0.995 \\
        40\% &  0.958 &  0.976 &  1.000 \\
        50\% &  0.980 &  0.991 &  1.000 \\
        \hline
        \end{tabular}
        }
    
    \caption{A fraction of instances that would be standardly selected on the current AL iteration, contained in top-k\% uncertain instances according to the acquisition model on the previous iteration (AG News corpus).}
    \label{tab:percents1}
    \vspace{-0.4cm}
\end{table}

\begin{table*}[ht]
\footnotesize
\centering
\scalebox{1.0}{\begin{tabular}{cllllll}
\hline
&{} &             \textbf{ELECTRA} & \textbf{BERT} & \textbf{DistilBERT} & \begin{tabular}[c]{@{}l@{}}\textbf{ELECTRA}\\ \textbf{with UPS (ours)}\end{tabular} & \begin{tabular}[c]{@{}l@{}}\textbf{DistilBERT}\\ \textbf{with UPS (ours)}\end{tabular} \\

\hline
\multirow{3}{*}{\rotatebox{90}{Iter. 2}} & Train             &      $176.3\pm 1.4$ &      $174.8\pm 1.4$ &      $87.4\pm 0.8$ &          $178.0\pm 1.4$ &              $87.9\pm 0.5$ \\
& Inference         &      $622.2\pm 9.4$ &      $623.8\pm 7.5$ &    $481.8\pm 17.2$ &         $630.9\pm 12.3$ &            $483.2\pm 23.0$ \\
& Overall           &      $798.6\pm 9.6$ &      $798.6\pm 8.4$ &    $569.2\pm 17.5$ &         $808.8\pm 12.8$ &            $571.1\pm 22.6$ \\
\hline
\multirow{3}{*}{\rotatebox{90}{Iter. 6}} & Train             &      $342.8\pm 5.7$ &      $339.9\pm 4.2$ &     $174.1\pm 2.9$ &          $342.2\pm 5.3$ &             $173.0\pm 1.4$ \\
& Inference         &     $600.5\pm 10.4$ &      $596.4\pm 6.6$ &     $455.1\pm 8.9$ &           \textbf{58.9}$\pm $\textbf{3.3} &              \textbf{50.0}$\pm $\textbf{6.4} \\
& Overall           &     $943.4\pm 15.9$ &      $936.3\pm 8.8$ &     $629.1\pm 9.7$ &          \textbf{401.1}$\pm $\textbf{3.4} &             \textbf{222.9}$\pm $\textbf{5.9} \\
\hline
\multirow{3}{*}{\rotatebox{90}{Iter. 10}} & Train             &      $504.6\pm 6.3$ &      $498.8\pm 3.9$ &     $257.5\pm 3.9$ &          $502.7\pm 6.0$ &             $255.1\pm 3.4$ \\
& Inference         &      $573.0\pm 6.9$ &      $577.5\pm 7.7$ &     $434.6\pm 4.6$ &           \textbf{55.5}$\pm $\textbf{2.9} &              \textbf{42.6}$\pm $\textbf{7.1} \\
& Overall           &    $1077.6\pm 13.1$ &    $1076.4\pm 10.9$ &     $692.1\pm 5.5$ &          \textbf{558.2}$\pm $\textbf{4.4} &            \textbf{297.7}$\pm $\textbf{10.3} \\
\hline
\multirow{3}{*}{\rotatebox{90}{Iter. 15}} & Train             &      $701.9\pm 7.2$ &     $714.9\pm 20.5$ &     $358.3\pm 3.0$ &         $704.8\pm 11.7$ &             $359.3\pm 5.4$ \\
& Inference         &      $548.6\pm 9.2$ &      $541.0\pm 5.0$ &    $415.9\pm 10.2$ &           \textbf{59.4}$\pm $\textbf{3.1} &              \textbf{39.3}$\pm $\textbf{2.6} \\
& Overall           &    $1250.5\pm 16.0$ &    $1255.9\pm 18.4$ &    $774.2\pm 10.8$ &         \textbf{764.2}$\pm $\textbf{10.6} &             \textbf{398.6}$\pm $\textbf{6.8} \\
\hline
                      \multicolumn{2}{l}{Overall train}     &    $6323.7\pm 72.1$ &    $6294.8\pm 73.7$ &   $3215.1\pm 38.5$ &        $6333.3\pm 92.8$ &           $3204.5\pm 32.5$ \\
\multicolumn{2}{l}{Overall inference} &   $8799.2\pm 150.7$ &   $8787.5\pm 102.7$ &   $6682.1\pm 96.2$ &        \textbf{3110.9}$\pm $\textbf{85.3} &           \textbf{2332.2}$\pm $\textbf{86.2} \\
\multicolumn{2}{l}{Overall}           &  $15122.9\pm 213.4$ &  $15082.2\pm 141.1$ &  $9897.1\pm 112.8$ &       \textbf{9444.2}$\pm $\textbf{113.6} &          \textbf{5536.7}$\pm $\textbf{100.8} \\
 \hline
\end{tabular}}

\caption{Duration of training and inference steps of AL iterations in seconds on AG News. Hardware configuration: 2 Intel Xeon Platinum 8168, 2.7 GHz, 24 cores CPU; NVIDIA Tesla v100 GPU, 32 Gb of VRAM.
}
\label{tab:time_agnews}
\end{table*}

\begin{figure}[t]
    \vspace{-0.1cm}
	\center{\includegraphics[width=1.\linewidth]{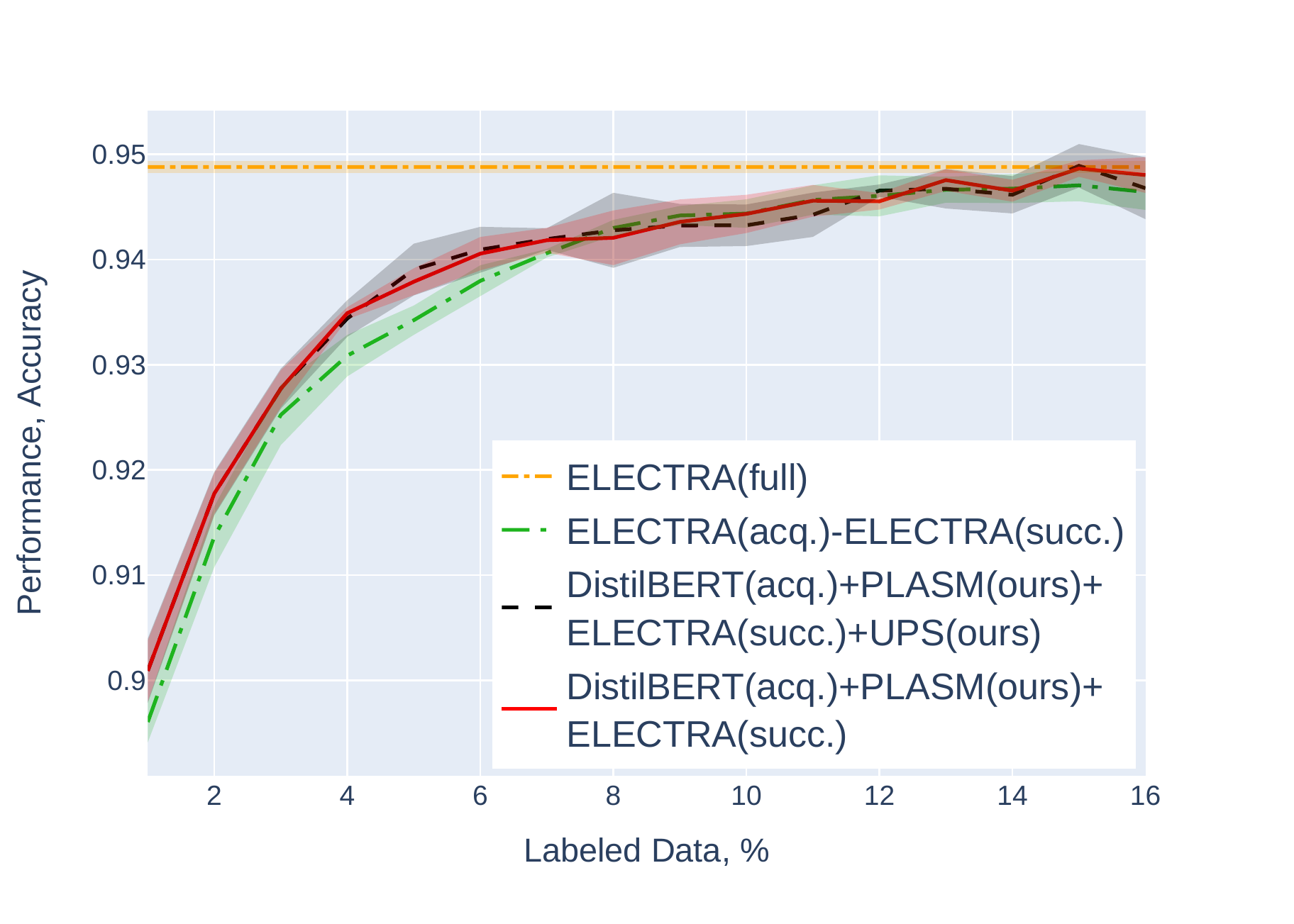}}
	\vspace{-0.7cm}
	\caption{The performance of UPS with PLASM (BERT is a pseudo-labeling model) on AG News compared with baselines ($\gamma = 0.1$, $T=0.01$).}
	\label{fig:ups_agnews}
	\vspace{-0.1cm}
	
\end{figure}

\subsubsection{Unlabeled Pool Subsampling}

Table~\ref{tab:time_agnews} compares the duration of AL iterations on the  AG News dataset, including the duration of the acquisition model training step and the duration of inference on instances from the unlabeled pool. 
We can see that the inference step is very time-consuming, especially on early iterations, and takes more than half of the time required for performing an AL iteration. Therefore, we claim that in such cases, it is more important to accelerate the inference step rather than the training step as it was done in previous work \cite{shen2018}.

To justify our approach to accelerating the inference step, we show that many unlabeled instances  have similar uncertainty estimates across different AL iterations. Table~\ref{tab:percents1} presents the fraction of instances, which would be standardly queried on the current iteration if we selected them from the whole unlabeled pool that are contained in k-\% of most uncertain instances, according to the acquisition model built on the previous AL iteration. For example, we observe that 50\% of the most uncertain instances according to the model trained on the first iteration contain more than 99\% of instances from the ``standard query'' on the second iteration, and 30\% contains almost 95\% of instances from the ``standard query''. Later iterations have even a better trade-off. Thereby, it is reasonable to avoid spending computational resources on instances that were most certain in previous iterations.


If we exclude a big part of the unlabeled pool from consideration during acquisition, the benefits of AL can potentially deteriorate. Results of experiments presented in Figure \ref{fig:ups_agnews} and Figures \ref{fig:ups_ag_and_conll}, \ref{fig:ups_conll} in Appendix~\ref{sec:add_exp_ups} show that the proposed UPS algorithm does not lead to the performance drop compared to the standard approach, in which we consider the whole unlabeled pool for instance selection. Meanwhile, the results of the ablation study in Figure \ref{fig:ups_random} in Appendix \ref{sec:add_exp_ups} demonstrate that the baseline, which randomly subsamples the
unlabeled dataset, has a performance drop compared to UPS.
In another ablation study, we set $T = 0$, which means that UPS just takes a fraction of the most uncertain instances (Figure \ref{fig:ups_T_0}). On some iterations, 
this results in a slight reduction of performance.   

From Table~\ref{tab:time_agnews}, we can see that UPS accelerates the query process up to 10 times. The corresponding results for CoNLL-2003 are presented in Table~\ref{tab:time_conll} in Appendix~\ref{sec:add_exp_ups}. Overall, applying both PLASM and UPS algorithms on AG News reduces the duration of AL iterations by more than 60\% compared with the standard approach. 
We can also tune the hyperparameters $\gamma$ and $T$ to reduce duration further in exchange for slightly worse scores. 



\section{Conclusion}\label{sec:concl}

We investigated several obstacles to deploying AL in practice and proposed two algorithms that help to overcome them. In particular, we considered the acquisition-successor mismatch problem revealed by \newcite{lowell2019practical}, as well as the problem related to the excessive duration of AL iterations with uncertainty-based query strategies and deep learning models. We demonstrate that the proposed PLASM algorithm helps to deal with both of these issues: it removes the constraint on the type of the successor model trained on the data labeled with AL and allows the use of lightweight acquisition models that have good training and inference performance, as well as a small memory footprint. The unlabeled pool subsampling algorithm helps to substantially decrease the inference time during AL without a loss in the quality of successor models. Together the PLASM and UPS algorithms help reduce the duration of an AL iteration by more than 60\%. We consider that the conducted empirical investigations and the proposed methods will help to increase the practicality of using deep AL in interactive annotation tools.

We note that applying PLASM requires some conditions to be met. Particularly, when a pseudo-labeling model is of considerably lower performance than a successor model, and filtering is not strict enough, training a successor directly on labeled instances acquired during AL with a different acquisition model may result in higher performance. Consequently, despite the pseudo-labeling model may be less expressive compared to the successor model, it should not be too ``weak''. In practice, we suggest comparing results obtained by the models trained with pseudo-labeling and without on a hold-out set and selecting the best model. 


There are still many issues that hinder the application of AL techniques. We consider that one of the most important obstacles is the necessity of hyperparameter optimization of deep learning models that can take a prohibitively long time to keep the annotation process interactive. We are looking forward to addressing this problem in future work.



\section*{Acknowledgements}

We thank anonymous reviewers for their insightful suggestions to improve this paper. The work was supported by a grant for research centers in the field of artificial intelligence (agreement identifier 000000D730321P5Q0002 dated November 2, 2021 No. 70-2021-00142 with ISP RAS).

\bibliography{anthology,custom}
\bibliographystyle{acl_natbib}


\clearpage
\appendix

\section{Dataset Statistics and Model Hyperparameters}
\label{app:hyp_data}

\vspace{-0.5cm}

\begin{table}[ht]
\label{tab:corp_stats1}
\centering

\caption{Dataset statistics. We provide a number of sentences/tokens for the training and test sets. k stands for a size of seeding datasets (\% of the training dataset) and a size of sets of instances selected for ``annotation'' on each iteration. C is a number of classes/entity types.}

\scalebox{0.78}{\begin{tabular}{lcccc}
\hline
\multicolumn{1}{l}{\textbf{Dataset}} & \multicolumn{1}{c}{\textbf{Train}} & \multicolumn{1}{c}{\textbf{Test}} & \multicolumn{1}{c}{\textbf{k}} & \multicolumn{1}{c}{\textbf{C}}\\ \hline
CoNLL-2003 & 15K/203.6K & 3.7K/46.4K & 2\% & 4(5)\\
OntoNotes 5.0 & 59.9K/1088.5K & 8.3K/152.7K & 2\% & 18\\ \hline
AG News & 120K/4541.7K & 7.6K/286.7K & 1\% & 4 \\ \hline
IMDb & 25K/5844.7K & 25K/5713.2K & 1\% & 2 \\ \hline
\end{tabular}}
\label{tab:corpora1}
\end{table}

\vspace{-0.4cm}

\begin{table}[ht]
\centering
\footnotesize
\caption{Hyperparameter values of Transformers. ``Sequence tagging'' incorporates CoNLL-2003 \& OntoNotes datasets, while ``Classification'' combines AG-News \& IMDB. The hyperparameters are chosen according to evaluation scores on the validation datasets when models are trained using the whole available training data on CoNLL-2003 for sequence tagging \& AG-News for classification.}

\begin{tabular}{lcc}
\hline
\multicolumn{1}{l}{\textbf{Hparam}} & \textbf{Sequence tagging} & \textbf{Classification}  \\
\hline
Number of epochs & 15 & 5 \\
Batch size & 16 & 16 \\
\multirowcell{Min. number of\\ training steps} & 1000 & 1000 \\
\begin{tabular}[l]{@{}l@{}}Max. sequence\\ length\end{tabular} & - & 256 \\
Optimizer & AdamW & AdamW \\
Learning rate & 5e-5 & 2e-5 \\
Weight decay & 0.01 & 0.01 \\
Gradient clipping & 1. & 1. \\
Scheduler & STLR & STLR \\
\% warm-up steps & 10 & 10 \\
\hline
\end{tabular}

\label{tab:hyp}
\end{table}

\begin{table}[ht]

\centering
\footnotesize

\caption{Transformers model checkpoints from HuggingFace repository \cite{Wolf2019HuggingFacesTS} \footnotemark .}

\scalebox{0.85}{\begin{tabular}{lccc}
\hline 
\textbf{Dataset} & \textbf{Model}  & \textbf{Checkpoint} & \textbf{\# Param.}\\
\hline
\multirow{4}{*}{\begin{tabular}[c]{@{}c@{}}AG-News /\\ IMDb\end{tabular}}
 & BERT  & bert-base-uncased & 110M \\
 & DistilBERT  & \begin{tabular}[c]{@{}c@{}}distilbert-base-
\\ uncased\end{tabular} & 67M\\
 & ELECTRA  & \begin{tabular}[c]{@{}c@{}}google/electra-\\ base-discriminator\end{tabular} & 110M \\
& DistilELECTRA  & \begin{tabular}[c]{@{}c@{}}lsanochkin/
\\ distilelectra-base\end{tabular} & 67M \\
& XLNet  & xlnet-base-cased & 117M \\
& RoBERTa  & roberta-base & 125M\\
& DistilRoBERTa  & distilroberta-base & 82M\\
\hline
\multirow{3}{*}{\begin{tabular}[c]{@{}c@{}}CoNLL-2003 /\\ OntoNotes 5.0\end{tabular}} & ELECTRA  & \begin{tabular}[c]{@{}c@{}}google/electra-\\ base-discriminator\end{tabular} & 110M\\
 & BERT  &bert-base-cased & 110M \\
 & DistilBERT  & \begin{tabular}[c]{@{}c@{}}distilbert-base-
\\ cased\end{tabular} & 67M
 \\\hline

\end{tabular}}
\label{tab:checkpoints}
\end{table}
\footnotetext[1]{\url{https://huggingface.co/models}}

\begin{table}[ht]
\centering
\footnotesize
\caption{Hyperparameter values of the CNN-BiLSTM-CRF model.}
\scalebox{0.96}{\begin{tabular}{@{}lc@{}}
\hline
\textbf{Hparam} & \textbf{CoNLL-2003} \\
\hline
\begin{tabular}[c]{@{}l@{}}Word embeddings\\ pre-trained model\end{tabular} & GloVe~\cite{pennington2014glove} \footnotemark \\
Word embedding dim. & 100 \\
Char embedding dim. & 30 \\
CNN dim. & 30 \\
CNN filters & {[}2, 3{]} \\
RNN num. layers & 1 \\
RNN hidden size & 200 \\
RNN word dropout prob. & 0.3 \\
RNN locked dropout prob. & 0.2 \\
Encoder dropout prob & 0.0 \\
Feed forward num. layers & 1 \\
Feed forward hidden size & 200 \\
Feed forward activation & Tanh \\
Feed forward dropout prob. & 0.0 \\
Batch size & 32 \\
Learning rate & 0.015 \\
Momentum & 0.9 \\
Number of epochs & 50 \\
Optimizer & SGD \\
Gradients clipping & 5 \\
\hline
\end{tabular}}
\label{tab:bilstm_hp}
\end{table}
\footnotetext[2]{\url{https://flair.informatik.hu-berlin.de/resources/embeddings/token/glove.gensim}}

\begin{table}[!ht]
\centering
\footnotesize
\caption{Hyperparameter values of the CNN model for text classification on AG News.}
\scalebox{0.96}{\begin{tabular}{@{}lc@{}}
\hline
\textbf{Hparam} & \textbf{AG News} \\
\hline
\begin{tabular}[c]{@{}l@{}}Word embeddings\\ pre-trained model\end{tabular} & Word2Vec~\cite{neurips-word2vec} \footnotemark \\
Word embedding dim. & 300 \\
CNN dim. & 100 \\
CNN filters & {[}3, 4, 5{]} \\
Dropout prob. & 0.5 \\
Batch size & 128 \\
Learning rate & 0.001 \\
Momentum & 0.9 \\
Number of epochs & 20 \\
Optimizer & SGD \\
Gradients clipping & 1 \\
\hline
\end{tabular}}
\label{tab:cnn_hp}
\end{table}
\footnotetext[3]{\url{https://drive.google.com/file/d/0B7XkCwpI5KDYNlNUTTlSS21pQmM/edit?usp=sharing}}

\vspace{-0.3cm}
\subsection{Distillation Details for the Custom DistilELECTRA Model}
\label{app:distillation}

The DistilELECTRA model is distilled from the ELECTRA-base model. It has the same architecture, but half as many layers initialized by taking from the teacher one layer out of two. Distillation is performed on the AG News dataset using a linear combination  $L=L_{ce}+L_{mlm}+L_{cos}+L_{mse}$ of the following loss functions as a training objective: $L_{ce} = \sum_i t_i \cdot log(s_i)$ is a distillation loss of the student's probabilities $s_i$ over the soft target probabilities of the teacher $t_i$; $L_{mlm}$ is the student's self-supervised masked language modeling loss; $L_{cos}$ is the cosine embedding loss that aligns the directions of the student's and teacher's hidden state vectors; $L_{mse}$ is a mean squared error between student's and corresponding teacher's hidden states vectors. DistilELECTRA is trained  with the following hyperparameters: $50$ epochs, batch size $5$, $50$ gradient accumulation steps, AdamW optimizer with a learning rate $5e-4$, epsilon $1e-6$.



\clearpage
\onecolumn
\section{Additional Experimental Results with Acquisition-successor Mismatch} \label{app:add_exp_asm}

\begin{figure*}[!ht]
    \footnotesize
    
    \centering
    \begin{minipage}[ht]{0.49\linewidth}
    \center{\includegraphics[width=1\linewidth]{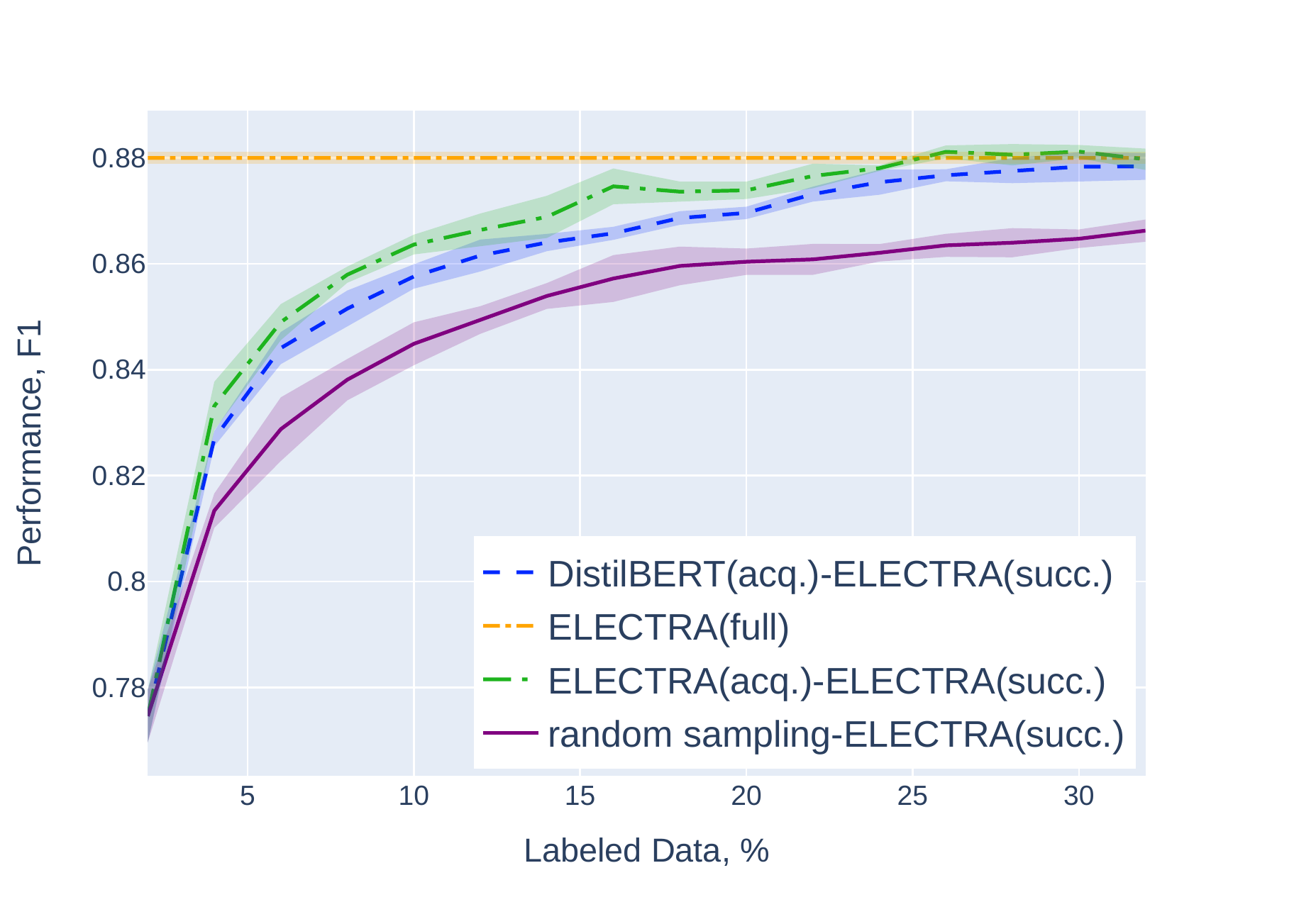} a) ELECTRA is a successor model. }
    \end{minipage}
    \hspace{0.1cm}
    \begin{minipage}[ht]{0.49\linewidth}
    \center{\includegraphics[width=1\linewidth]{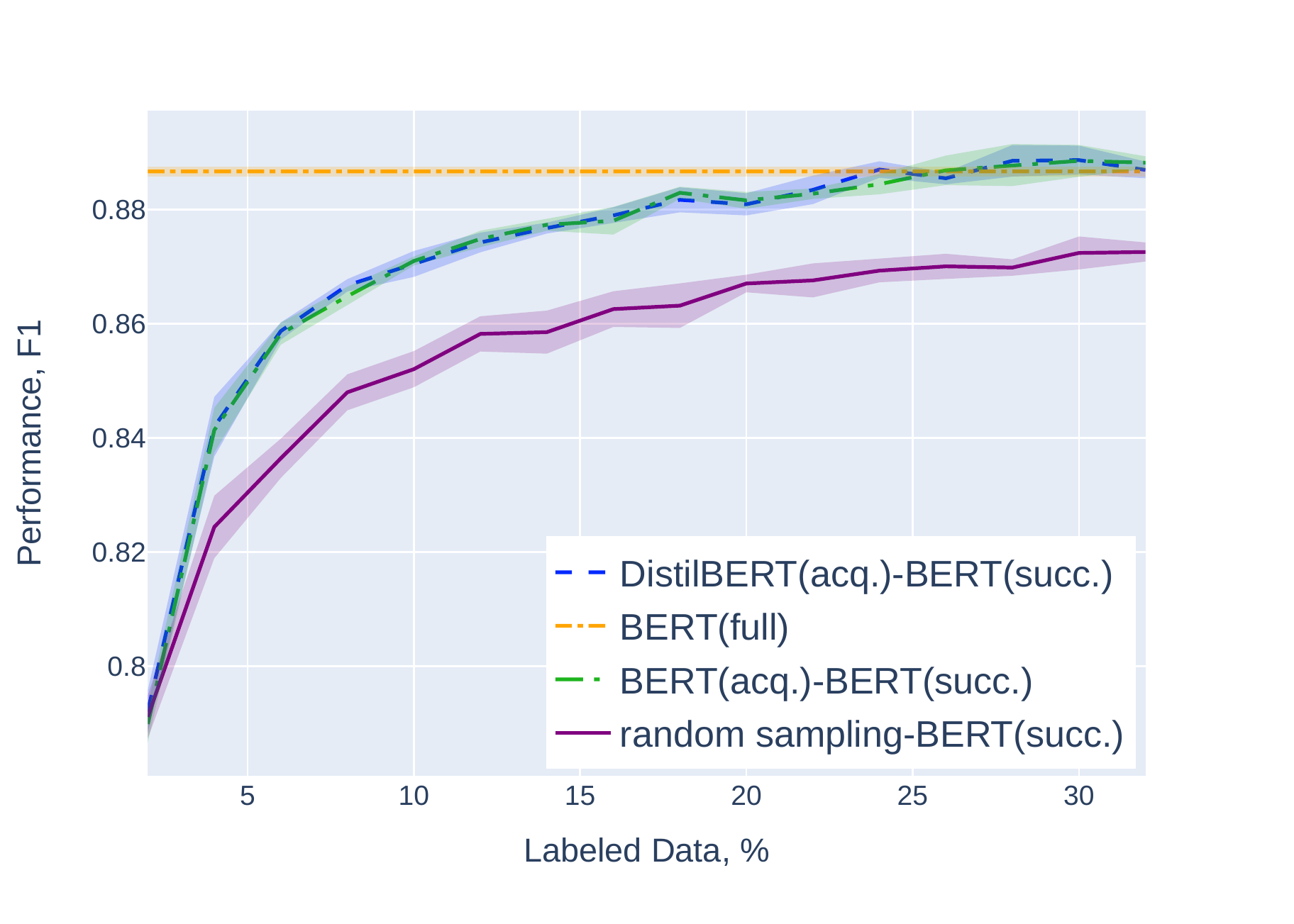} b) BERT is a successor model.}
    \end{minipage}
    
    \caption{AL experiments on OntoNotes, in which a successor model does not match an acquisition model (DistilBERT).}
    \label{fig:asm_onto}
\end{figure*}
\vspace{-0.5cm}



\begin{figure*}[!ht]
    \footnotesize
    
    \centering
    \begin{minipage}[h]{0.49\linewidth}
    \center{\includegraphics[width=1\linewidth]{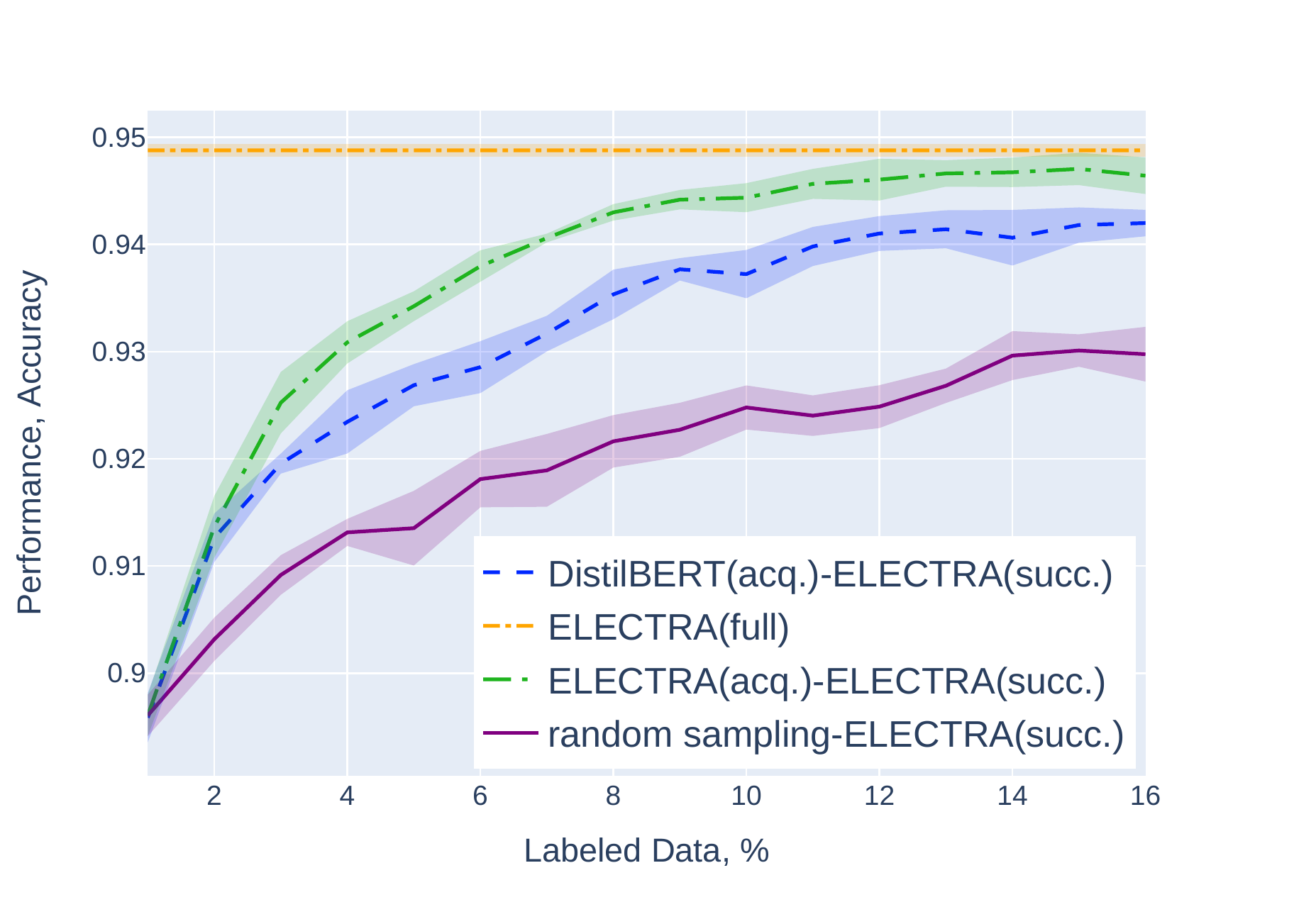} a) ELECTRA is a successor model. }
    \end{minipage}
    \hspace{0.1cm}
    \begin{minipage}[h]{0.49\linewidth}
    \center{\includegraphics[width=1\linewidth]{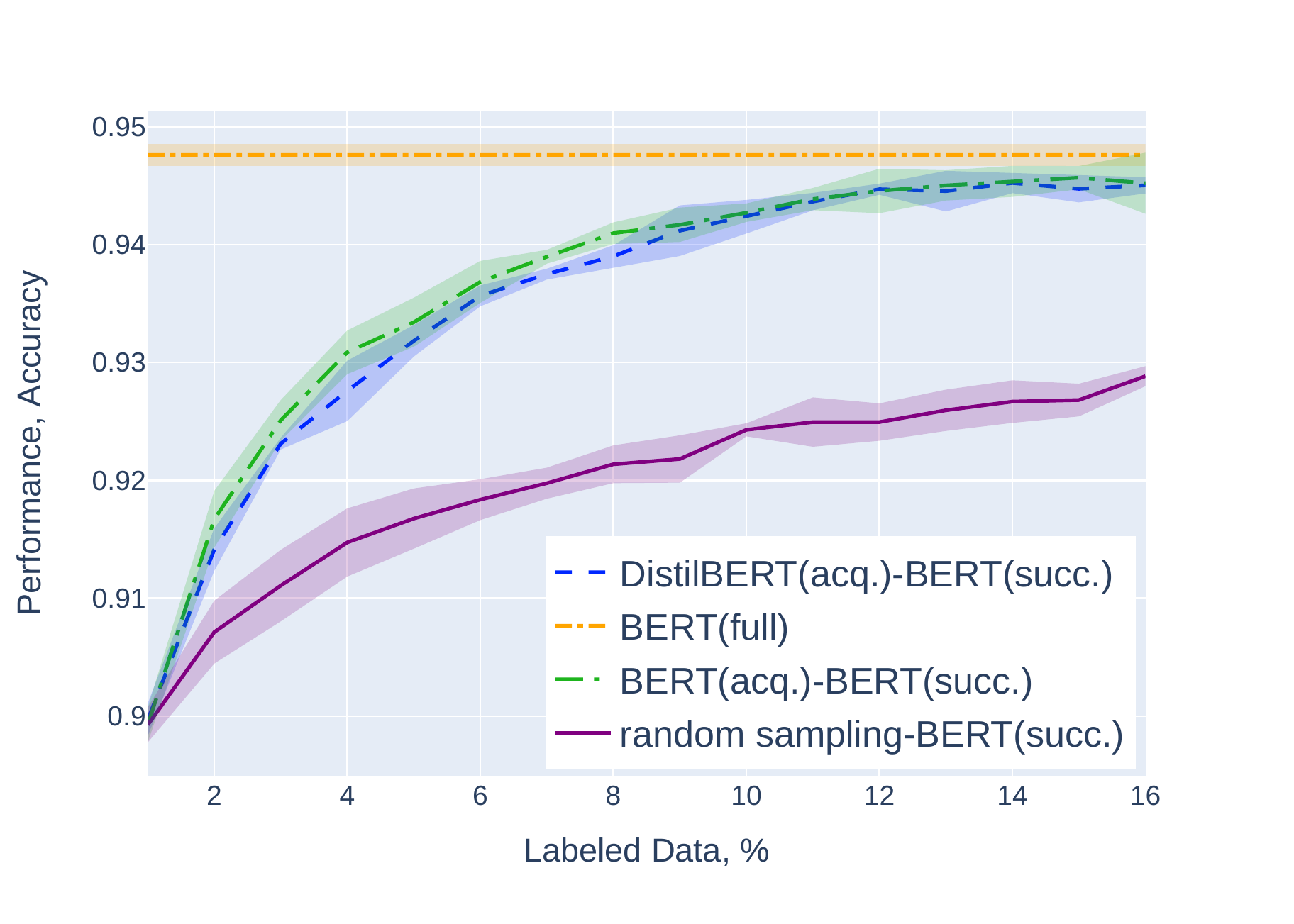} b) BERT is a successor model.}
    \end{minipage}
    
    \caption{AL experiments on AG News, in which a successor model does not match an acquisition model (DistilBERT).}
    \label{fig:asm_ag}
\end{figure*}
\vspace{-0.5cm}

\begin{figure*}[!ht]
    \footnotesize
    
    \centering
    \begin{minipage}[h]{0.49\linewidth}
    \center{\includegraphics[width=1\linewidth]{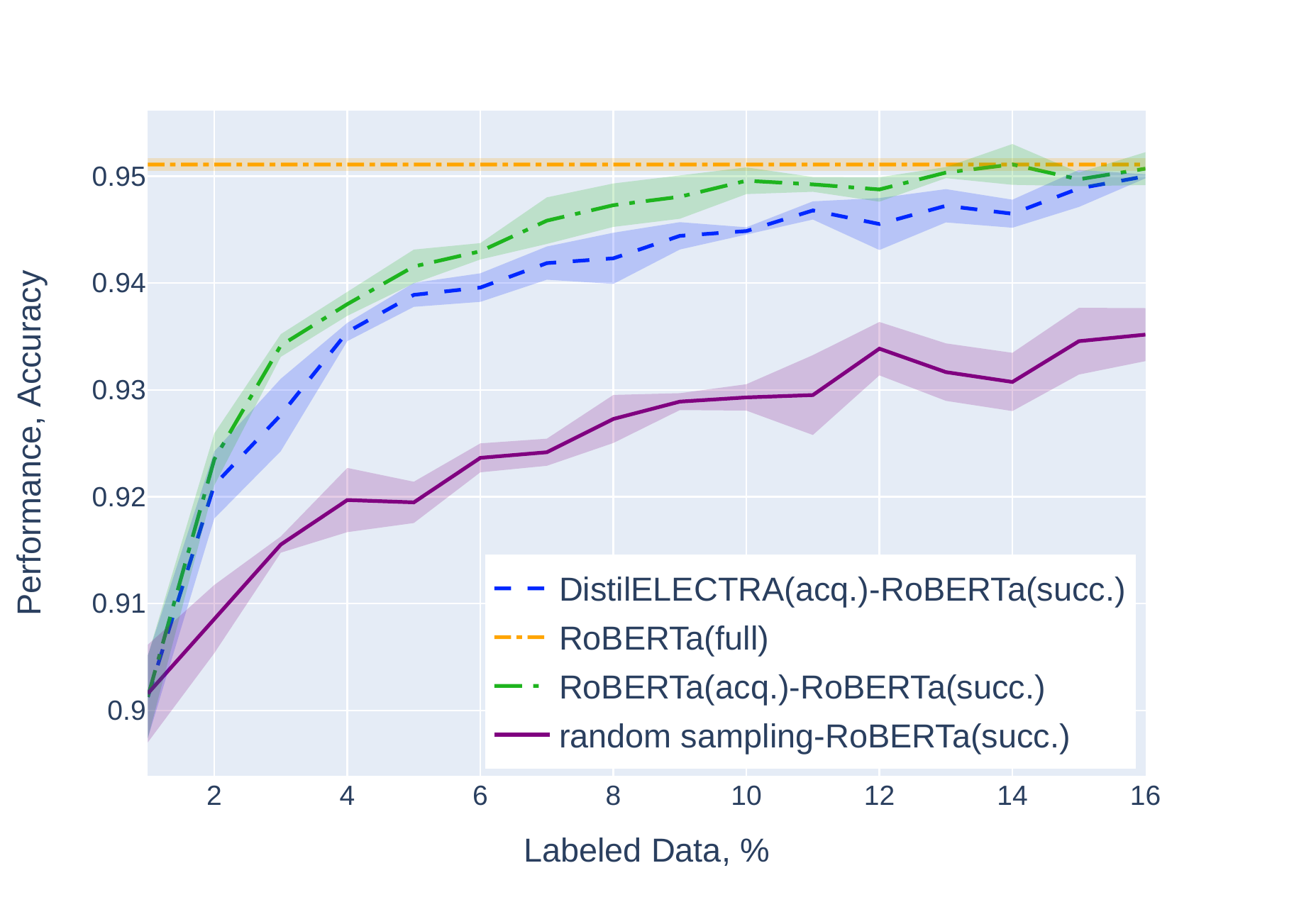} a) RoBERTa is a successor model. }
    \end{minipage}
    \hspace{0.1cm}
    \begin{minipage}[h]{0.49\linewidth}
    \center{\includegraphics[width=1\linewidth]{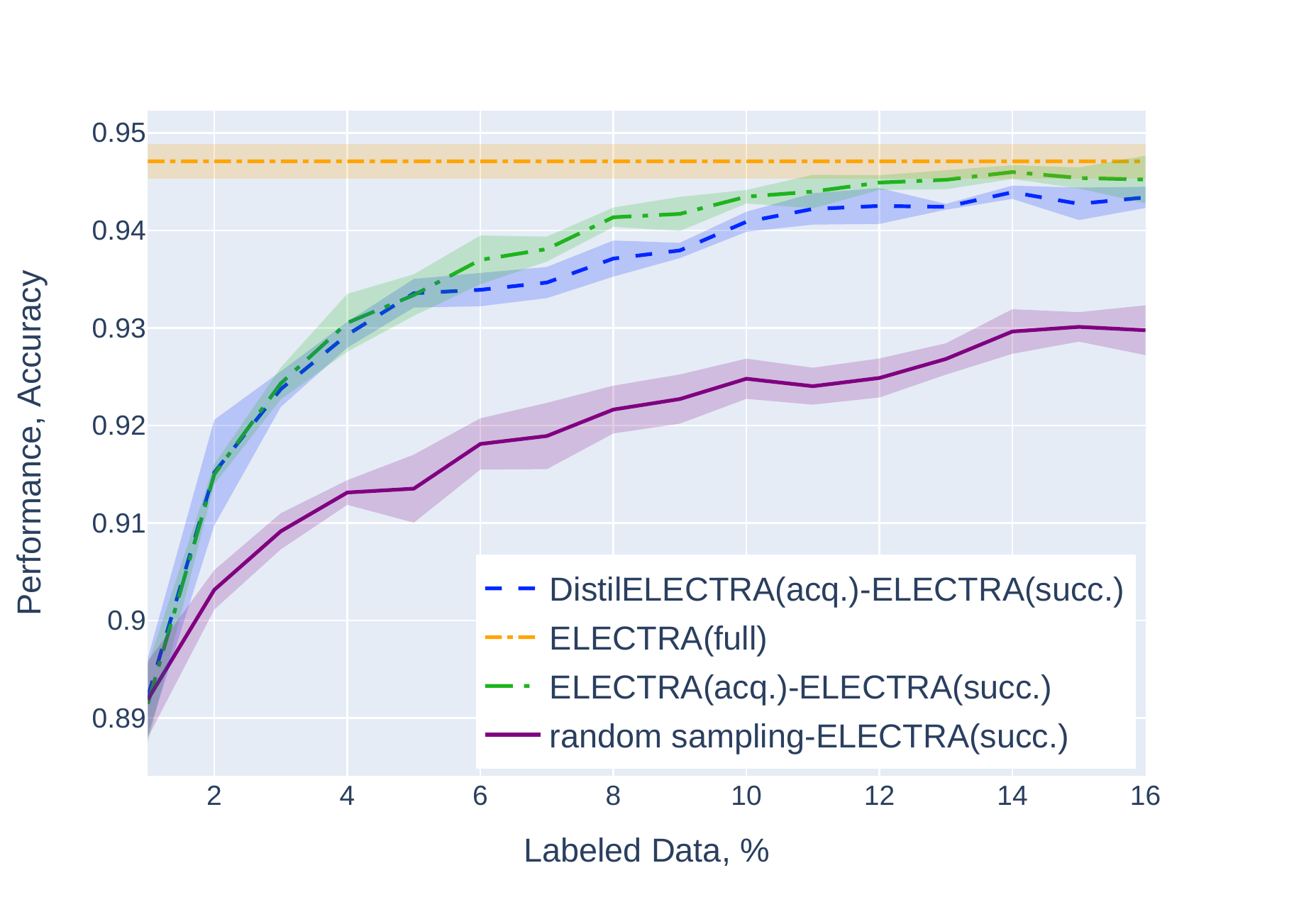} b) ELECTRA is a successor model.}
    \end{minipage}
    
    \caption{AL experiments on AG News, in which a successor model does not match an acquisition model (DistilELECTRA).}
    \label{fig:asm_ag_distilelectra}
\end{figure*}
\vspace{-0.5cm}

\begin{figure*}[!ht]
    \footnotesize
    
    \centering
    \begin{minipage}[h]{0.49\linewidth}
    \center{\includegraphics[width=1\linewidth]{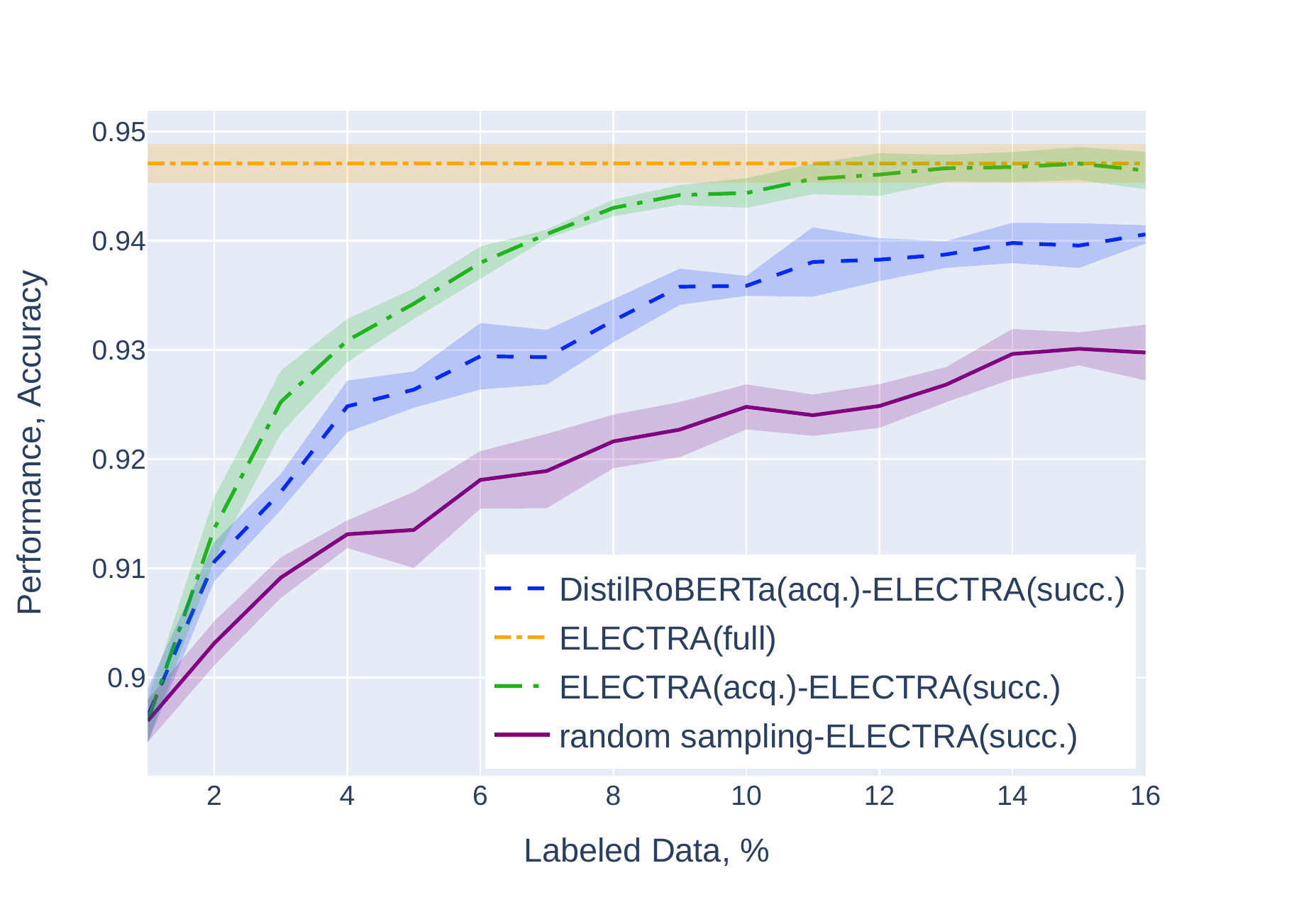} a) ELECTRA is a successor model. }
    \end{minipage}
    \hspace{0.1cm}
    \begin{minipage}[h]{0.49\linewidth}
    \center{\includegraphics[width=1\linewidth]{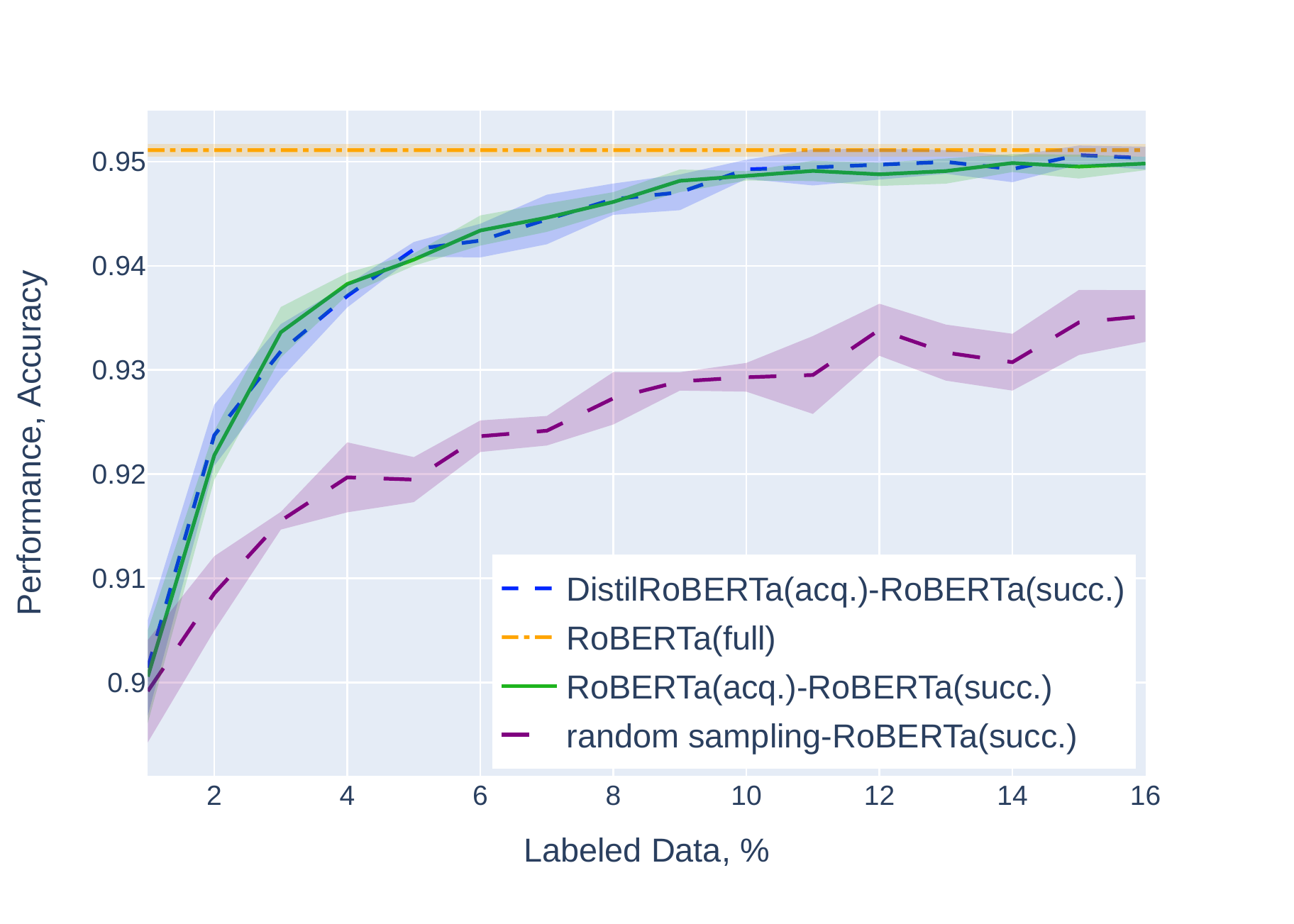} b) RoBERTa is a successor model.}
    \end{minipage}
    
    \caption{AL experiments on AG News, in which a successor model does not match an acquisition model (DistilRoBERTa).}
    \label{fig:asm_ag_distilroberta}
\end{figure*}
\vspace{-0.5cm}

\begin{figure*}[!ht]
    \footnotesize
    
    \centering
    \begin{minipage}[h]{0.49\linewidth}
    \center{\includegraphics[width=1\linewidth]{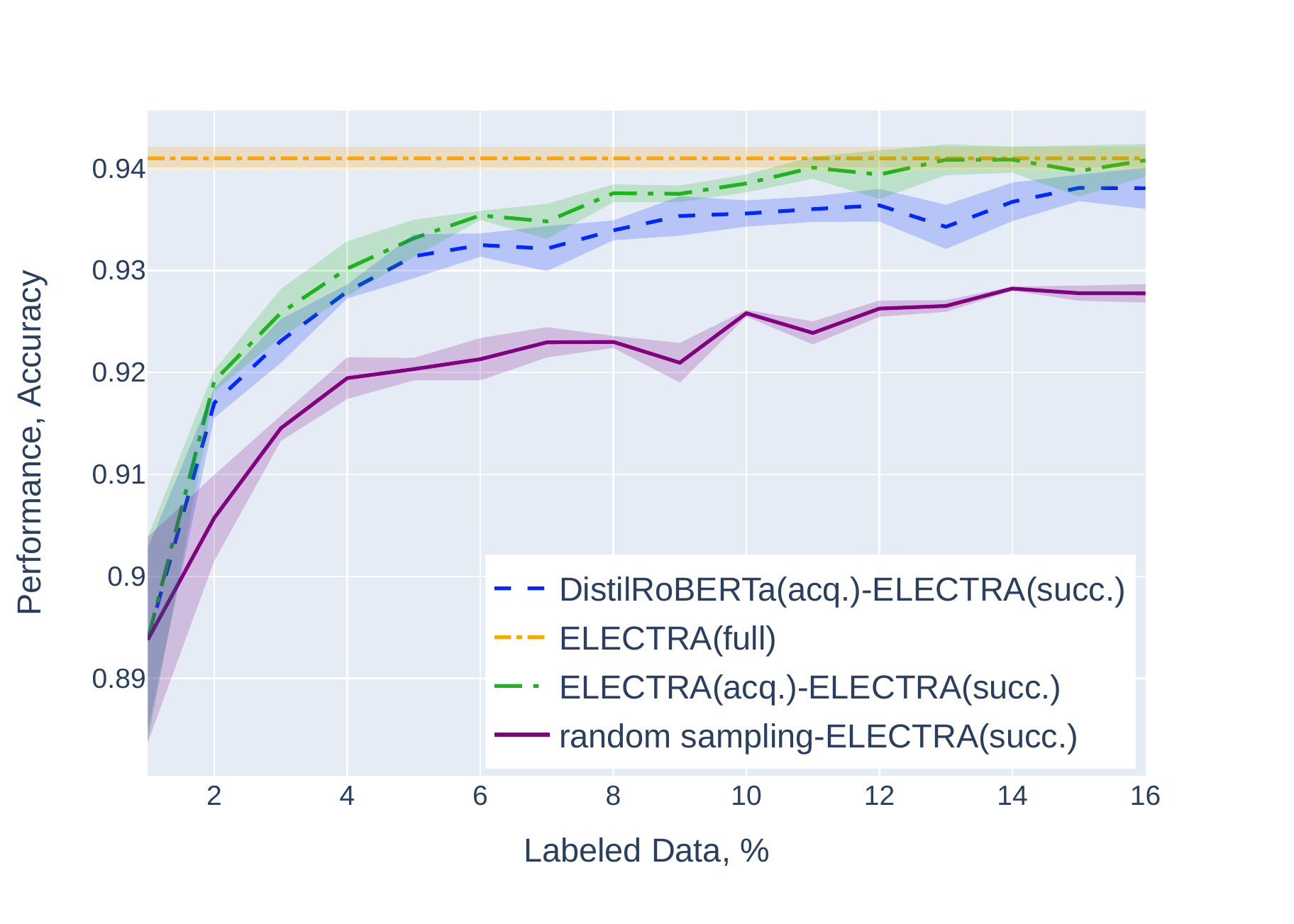} a) ELECTRA is a successor model. }
    \end{minipage}
    \hspace{0.1cm}
    \begin{minipage}[h]{0.49\linewidth}
    \center{\includegraphics[width=1\linewidth]{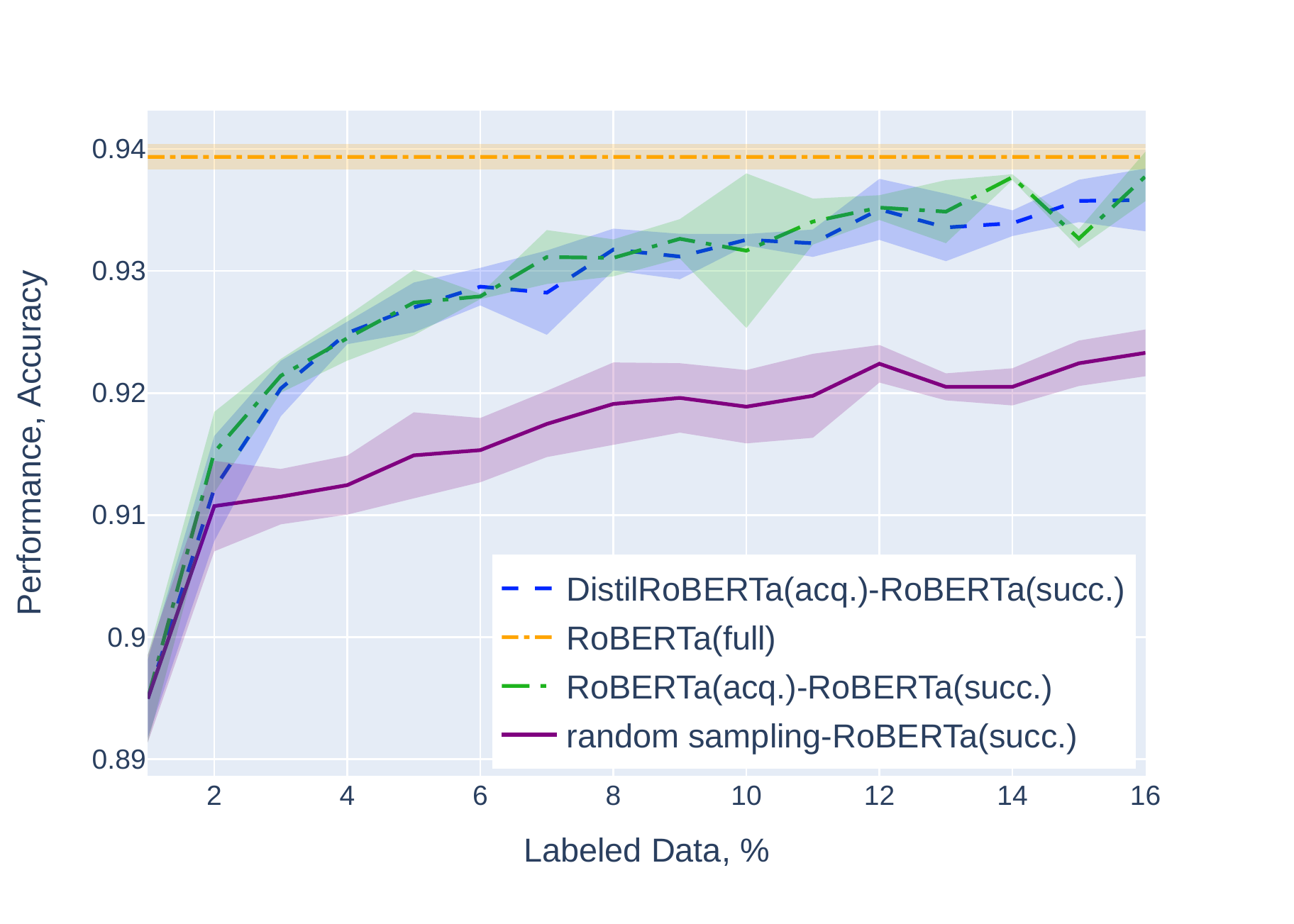} b) RoBERTa is a successor model.}
    \end{minipage}
    
    \caption{AL experiments on IMDb, in which a successor model does not match an acquisition model (DistilRoBERTa).}
    \label{fig:asm_imdb}
\end{figure*}
\vspace{-0.5cm}

\begin{figure}[!ht]
    \center{\includegraphics[width=0.5\linewidth]{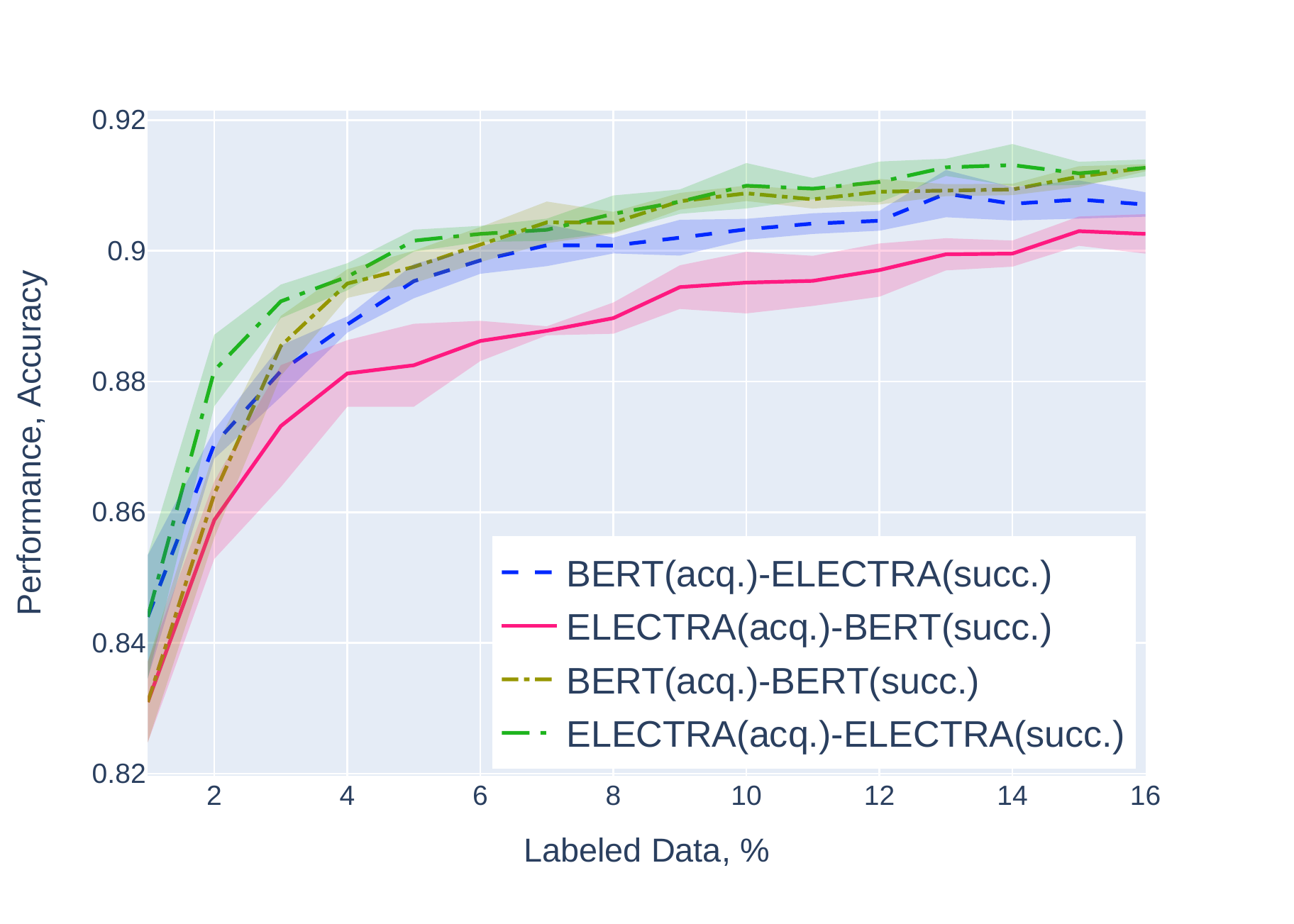}}
	\caption{AL experiments on CoNLL-2003, in which a successor model does not match an acquisition model. This experiment demonstrates that models with similar expressiveness and size (BERT and ELECTRA) cannot be used interchangeably for acquisition in AL.}
	
	\label{fig:conll_asm_bert_electra}
\end{figure}

\begin{figure}[!ht]
    \center{\includegraphics[width=0.5\linewidth]{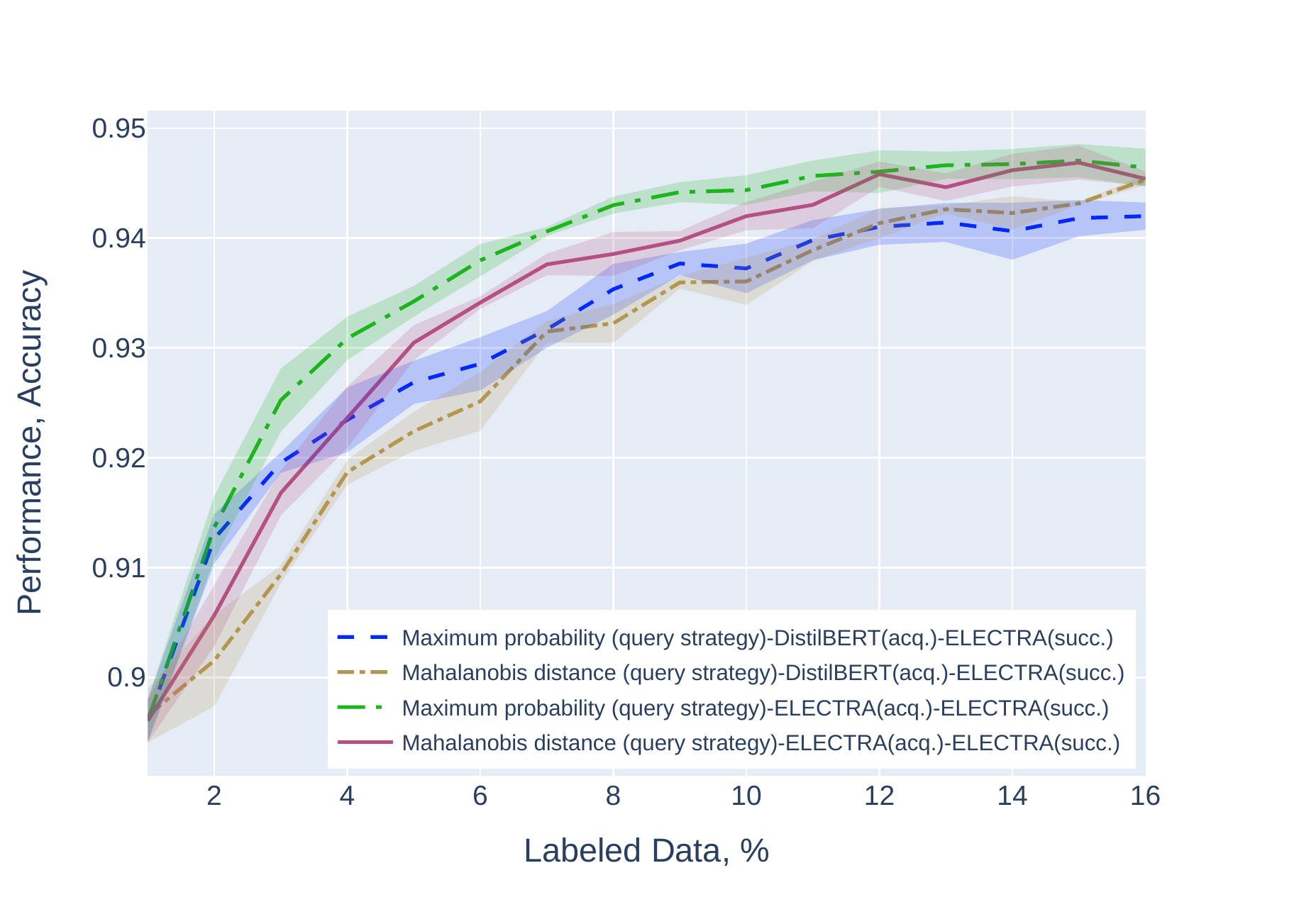}}
	\caption{AL experiments on AG News with Mahalanobis distance used as an uncertainty measure in a query strategy. This experiment demonstrates that the acquisition-successor mismatch problem also persists for this modern uncertainty estimation technique.}
	
	\label{fig:asm_mahalanobis}
\end{figure}




\clearpage
\vspace{1.5cm}
\section{Additional Experimental Results with PLASM} \label{sec:add_exp_plasm}

\begin{figure*}[!ht]
    \footnotesize
    
    \centering
    \begin{minipage}[h]{0.49\linewidth}
    \center{\includegraphics[width=1\linewidth]{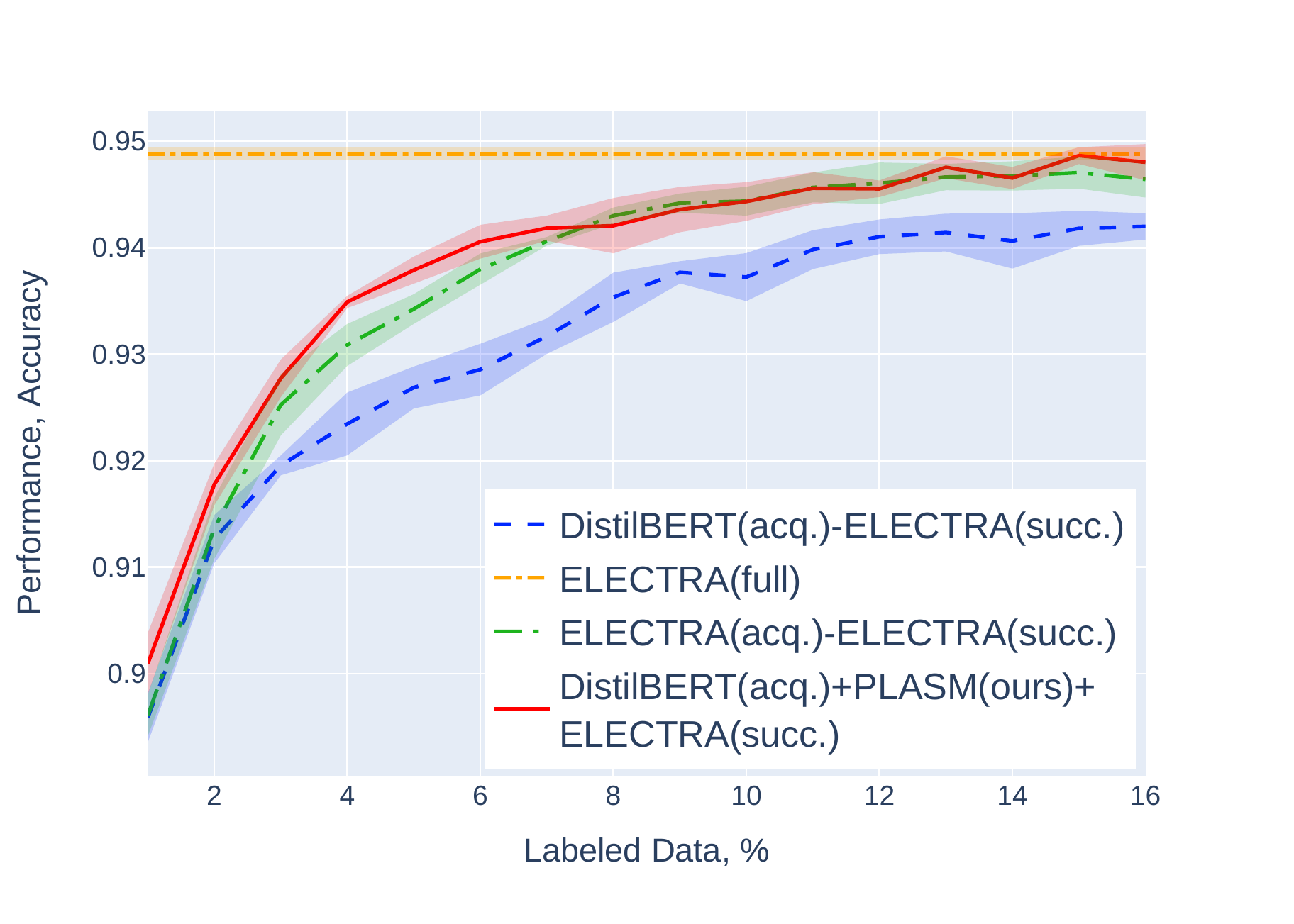} a) DistilBERT is an acquisition model, BERT is a pseudo-labeling model, ELECTRA is a successor model.}
    \end{minipage}
    \hspace{0.1cm}
    \begin{minipage}[h]{0.49\linewidth}
    \center{\includegraphics[width=1\linewidth]{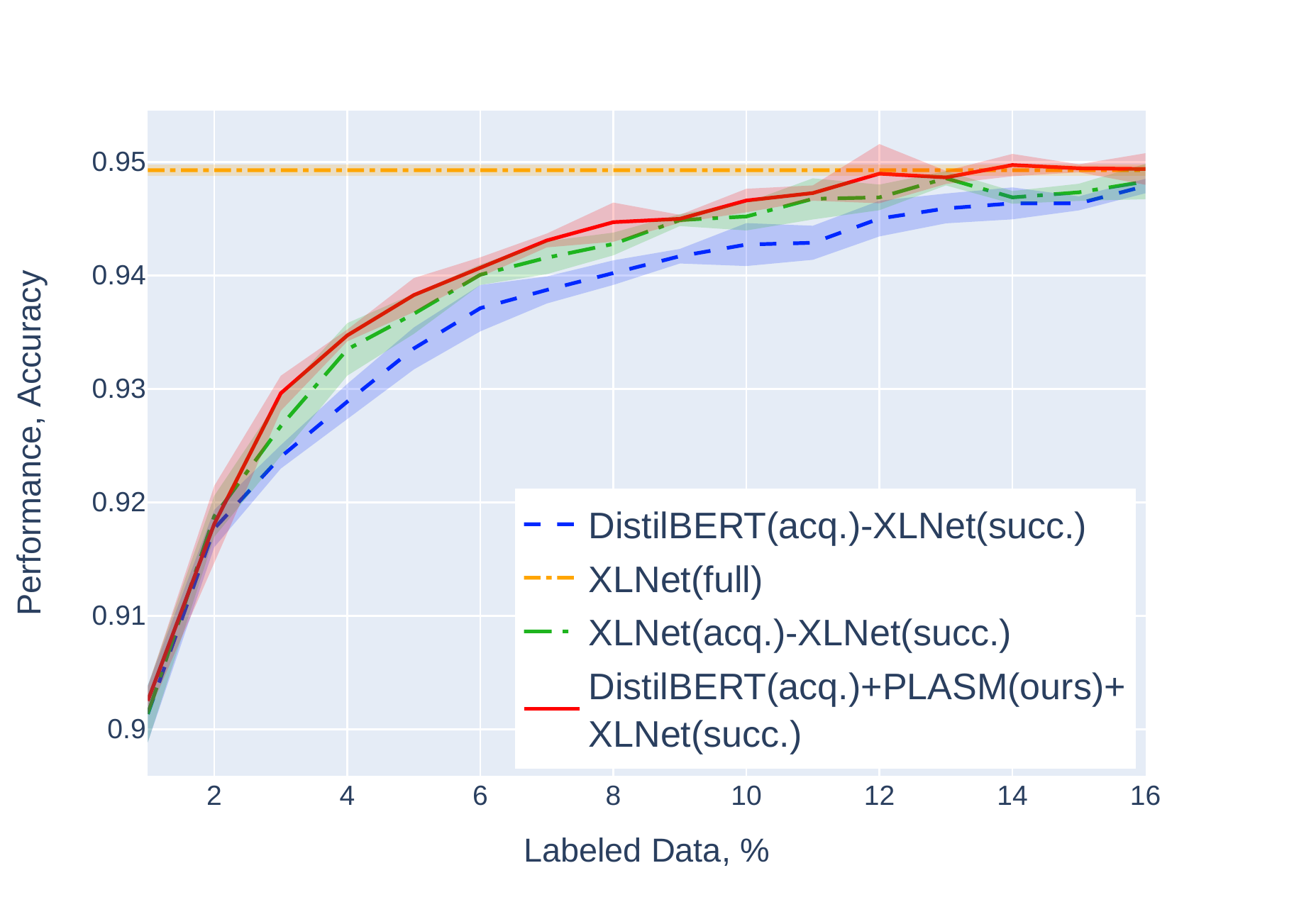} b) DistilBERT is an acquisition model, BERT is a pseudo-labeling model, XLNet is a successor model.}
    \end{minipage}

    \centering
    \begin{minipage}[h]{0.49\linewidth}
    \center{\includegraphics[width=1\linewidth]{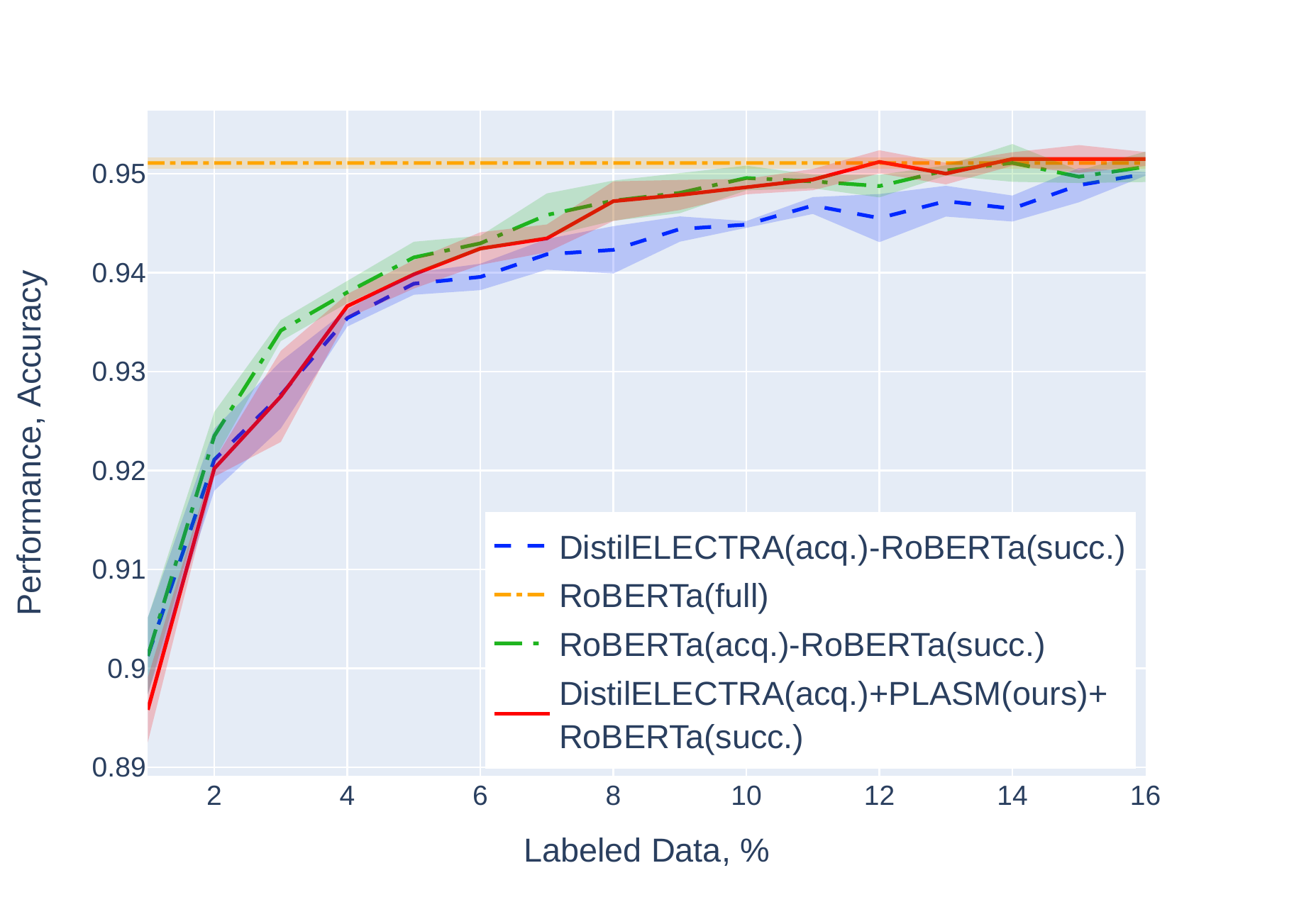} c) DistilELECTRA is an acquisition model, ELECTRA is a pseudo-labeling model, RoBERTa is a successor model.}
    \end{minipage}
    \hspace{0.1cm}
    \begin{minipage}[h]{0.49\linewidth}
    \center{\includegraphics[width=1\linewidth]{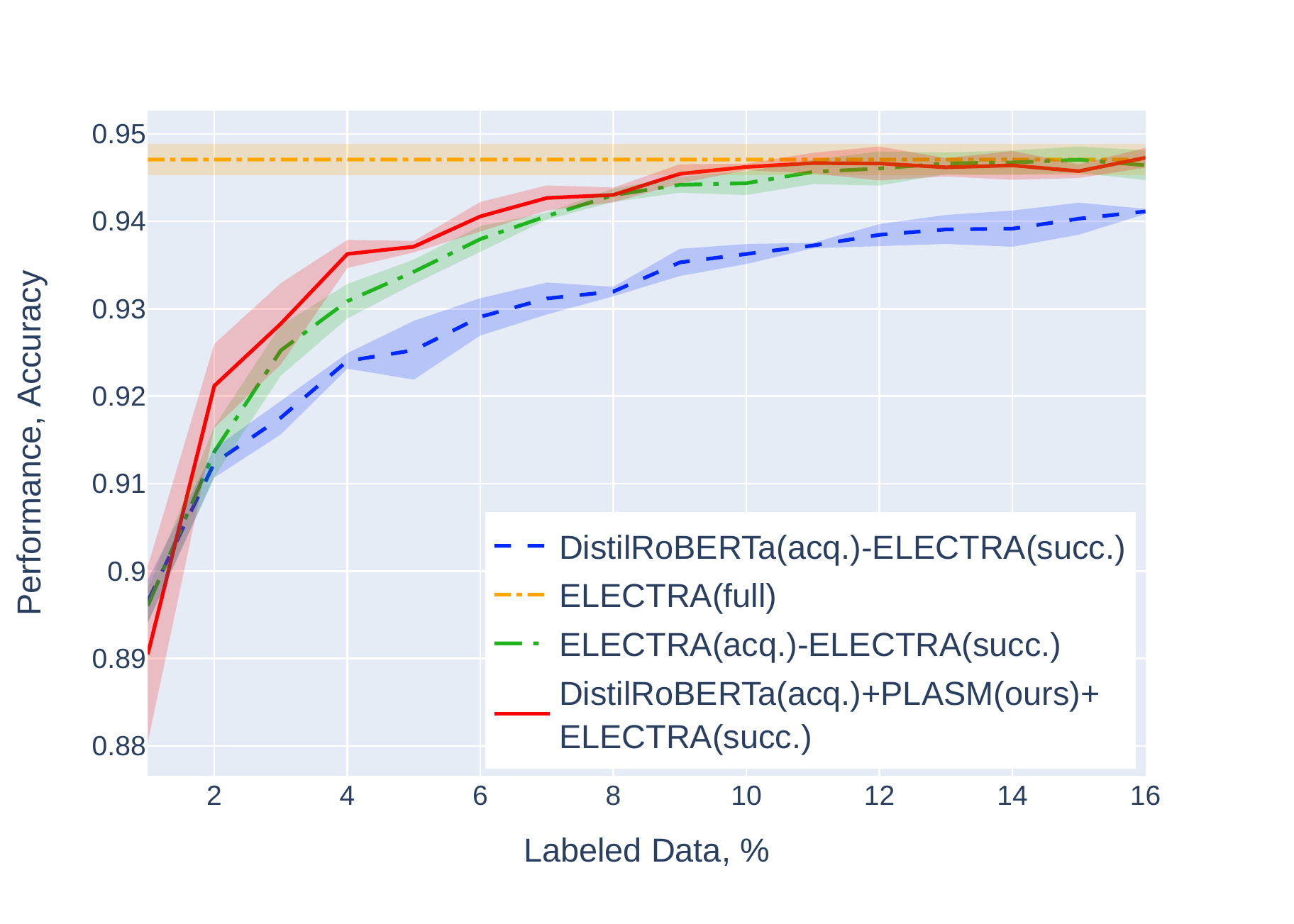} d) DistilRoBERTa as an acquisition model, RoBERTa as a pseudo-labeling model, ELECTRA is a successor model.}
    \end{minipage}
    
    \caption{The performance of PLASM compared with the standard approach to AL on AG News with various acquisition ~-- pseudo-labeling model pairs and successor models.}
    \label{fig:plasm_ag}
\end{figure*}

    
    
    

\begin{figure}[!ht]
    \center{\includegraphics[width=0.5\linewidth]{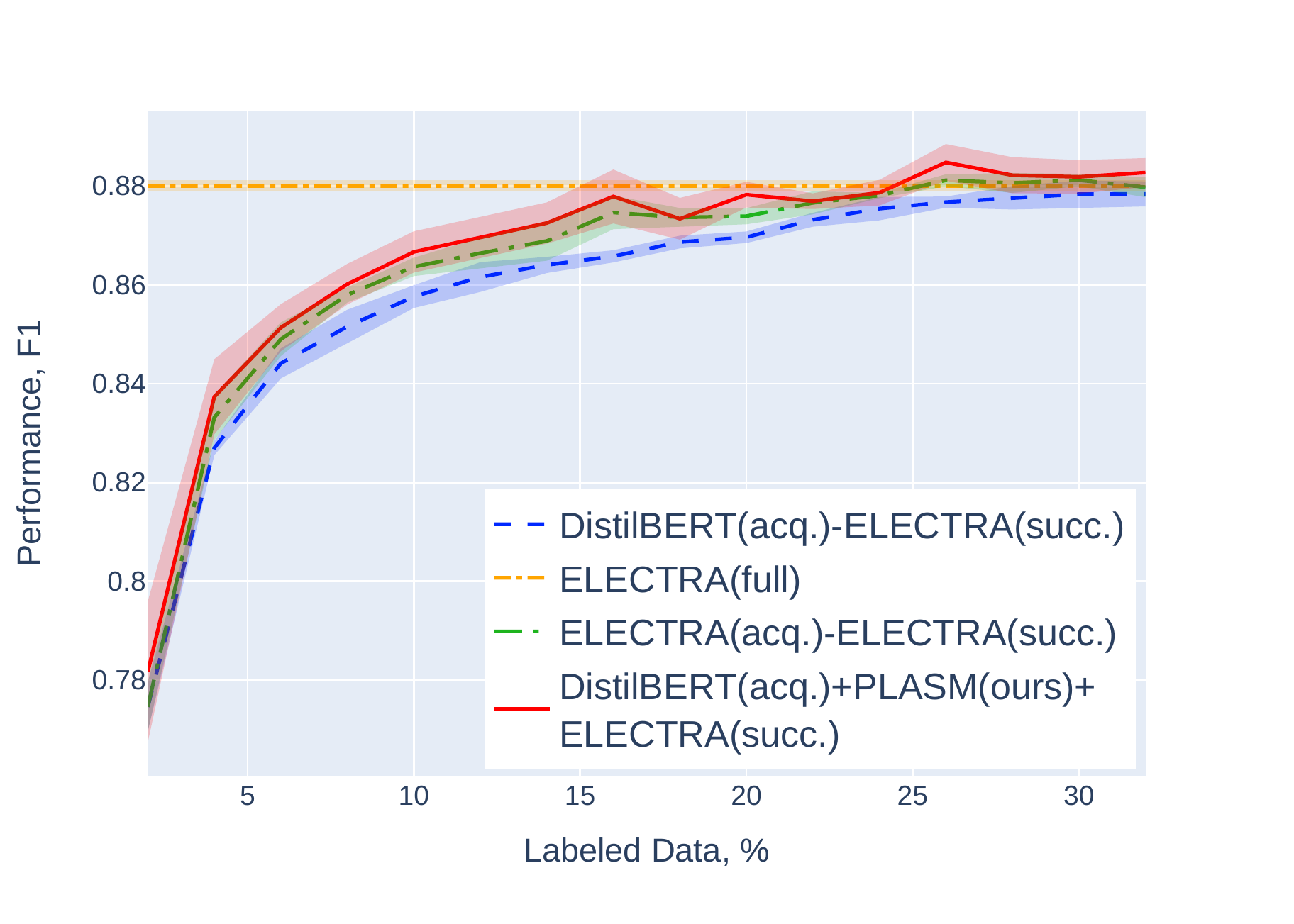}}
	\caption{The performance of PLASM (BERT is a pseudo-labeling model) compared with the standard approach to AL on OntoNotes.}
	\label{fig:plasm_onto}
\end{figure}

\begin{figure}[!ht]
    \center{\includegraphics[width=0.5\linewidth]{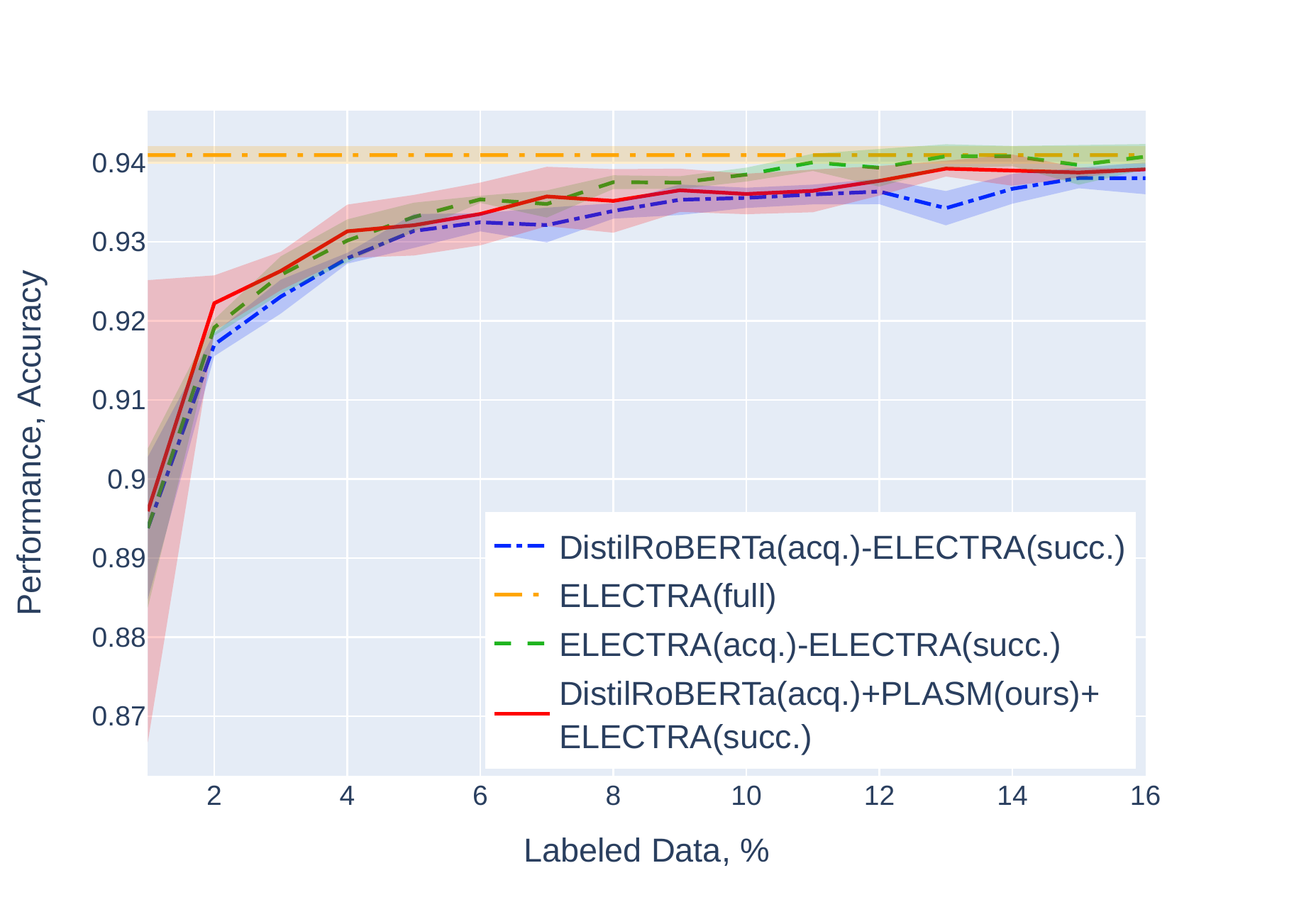}}
	\caption{The performance of PLASM (RoBERTa is a pseudo-labeling model) compared with the standard approach to AL on IMDb.}
	\label{fig:plasm_imdb}
\end{figure}

\begin{figure}[!ht]
    \center{\includegraphics[width=0.5\linewidth]{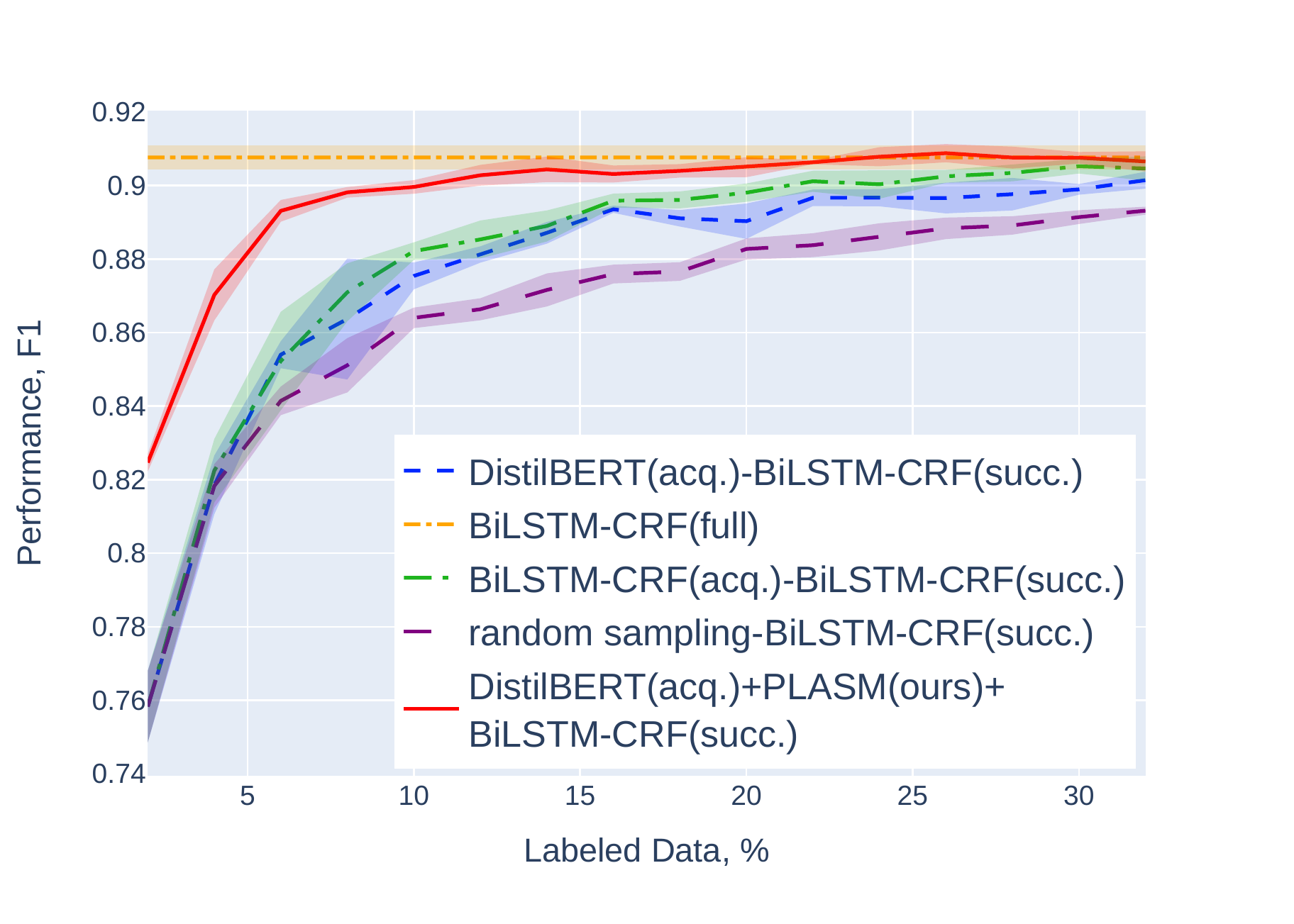}}
	\caption{Experiments with PLASM and standard approaches on CoNLL-2003, in which CNN-BiLSTM-CRF is used as a successor model. We can see that due to using PLASM and the expressiveness of the pseudo-labeling model (BERT), the successor achieves substantial improvements over the baseline.}
	
	\label{fig:conll_bilstm}
\end{figure}

\begin{figure}[!ht]
    \center{\includegraphics[width=0.5\linewidth]{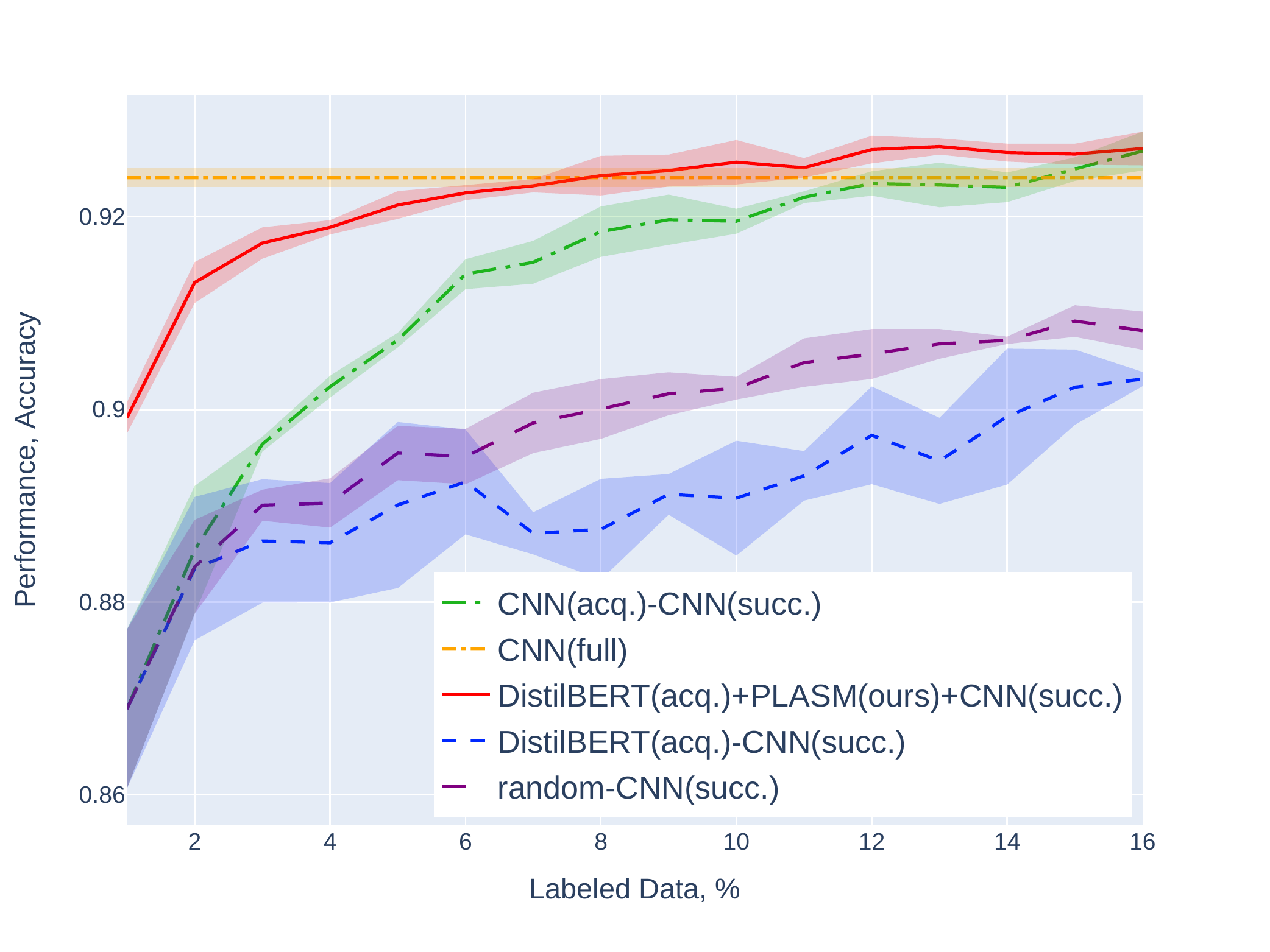}}
	\caption{Experiments with PLASM and standard approaches on AGNews, in which simple CNN is used as a successor model. We can see that due to using PLASM and the expressiveness of the pseudo-labeling model (BERT), the successor achieves substantial improvements over the baseline. We also note that using AL with DistilBERT as an acquisition model results in worse performance than using the baseline random sampling; this corresponds to findings of \cite{lowell2019practical}.}
	
	\label{fig:agnews_cnn}
\end{figure}

\begin{figure*}[!ht]
    \footnotesize
    
    \centering
    \begin{minipage}[ht]{0.49\linewidth}
    \center{\includegraphics[width=1\linewidth]{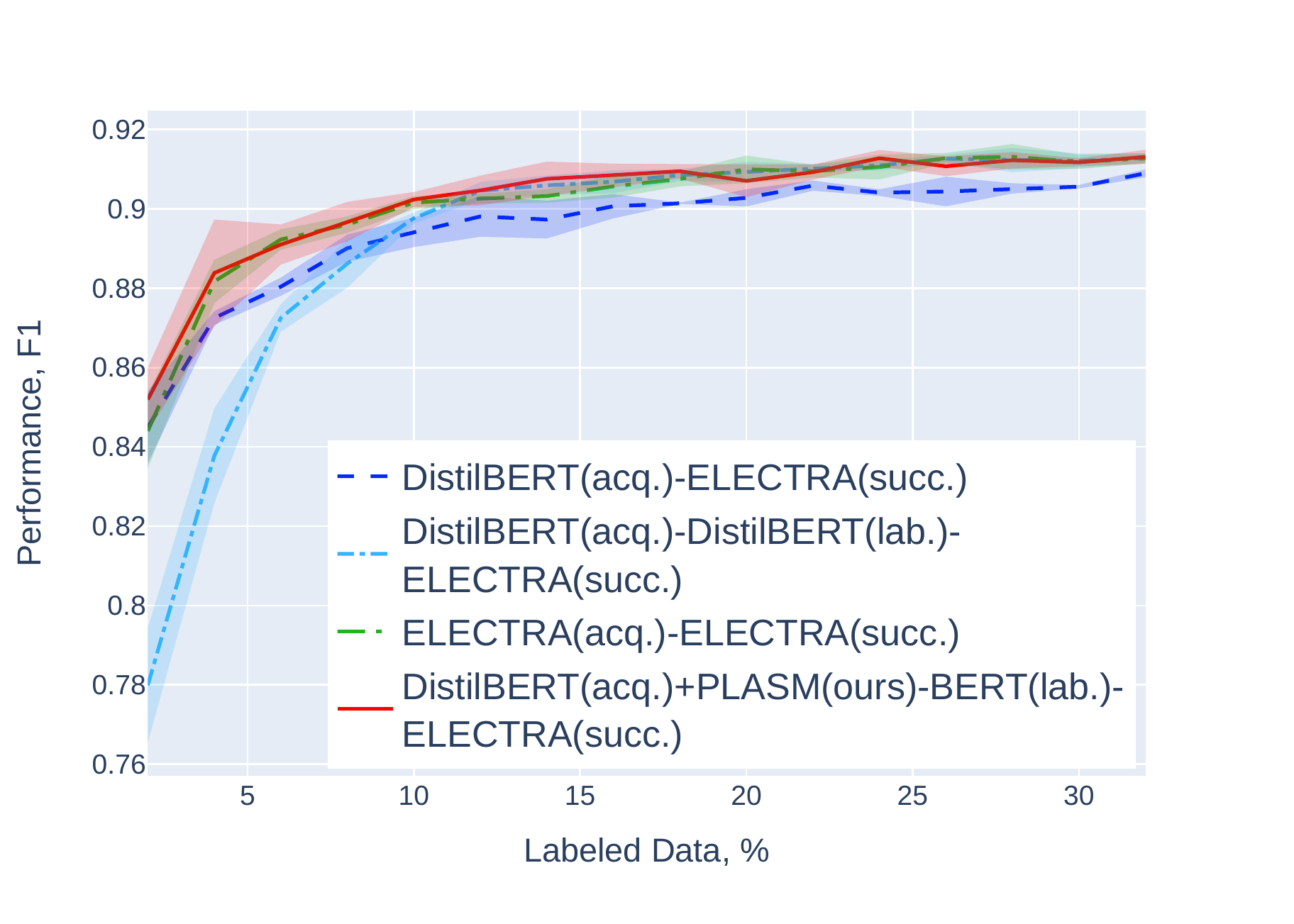} a) DistilBERT for pseudo-labeling. }
    \end{minipage}
    \hspace{0.1cm}
    \begin{minipage}[ht]{0.49\linewidth}
    \center{\includegraphics[width=1\linewidth]{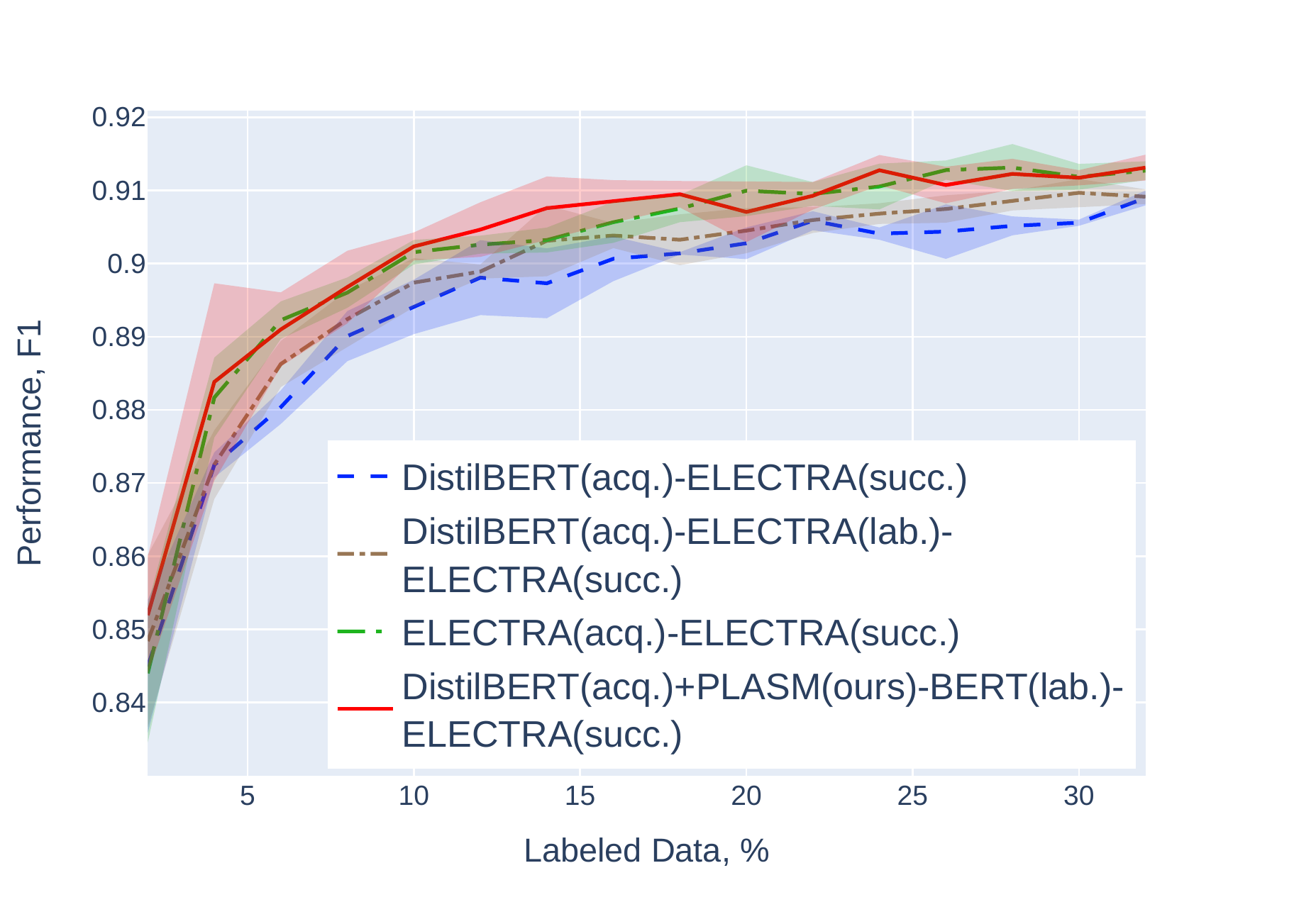} b) ELECTRA for pseudo-labeling.}
    \end{minipage}
    
    \caption{Ablation studies of PLASM on the CoNLL-2003 dataset, in which an inappropriate model is used for pseudo-labeling. }
    
    \label{fig:plasm_ablation}
\end{figure*}

\begin{figure*}[!ht]
    \footnotesize
    
    \centering
    \begin{minipage}[ht]{0.49\linewidth}
    
    \center{\includegraphics[width=1\linewidth]{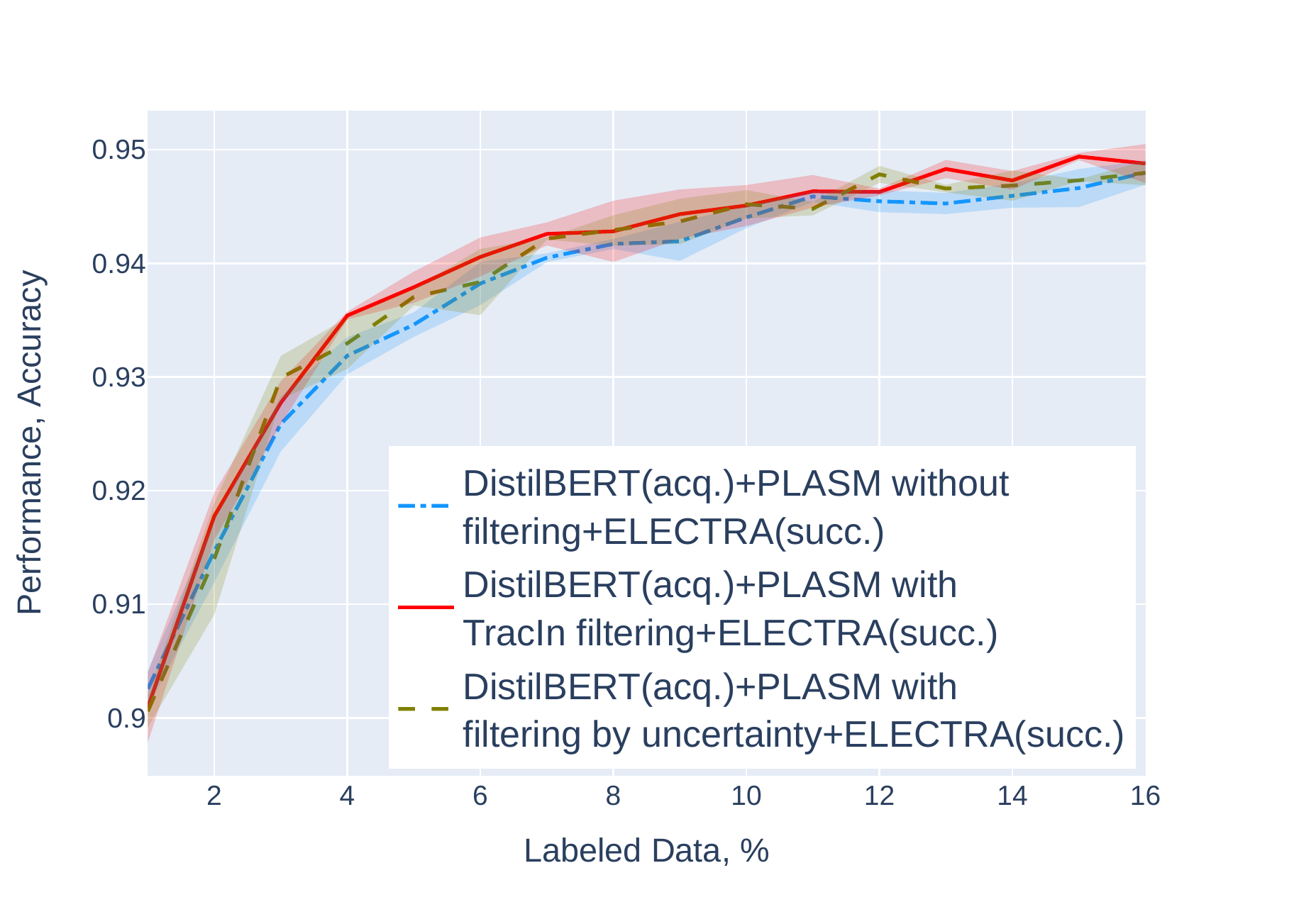} a) ELECTRA as a successor. }
    \end{minipage}
    \hspace{0.1cm}
    \begin{minipage}[ht]{0.49\linewidth}
    
    \center{\includegraphics[width=1\linewidth]{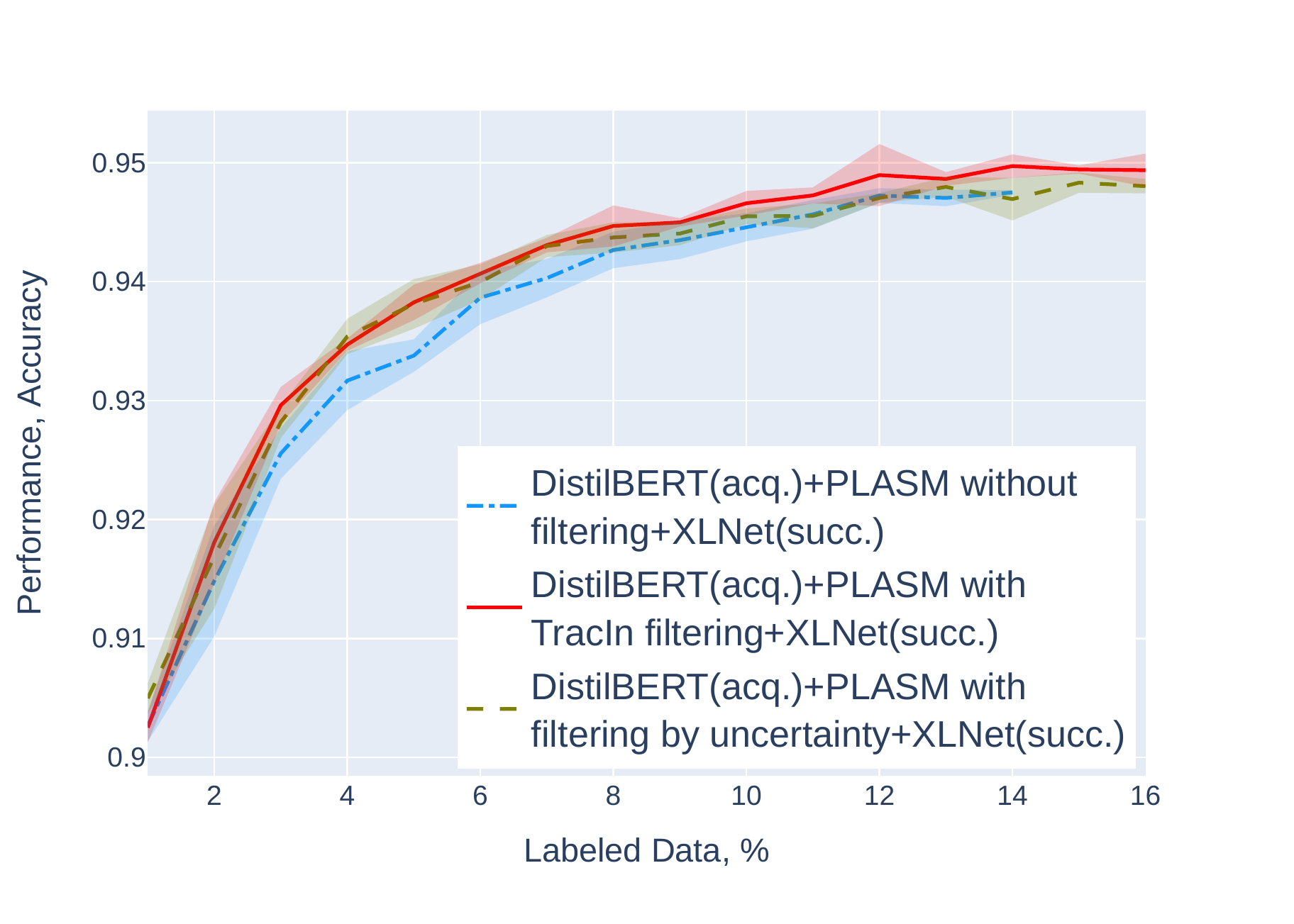} b) XLNet as a successor.}
    \end{minipage}
    
    \caption{Ablation studies of filtering methods in PLASM on the AG News dataset. }
    
    \label{fig:plasm_ablation_filtering}
    \vspace{1.5cm}
\end{figure*}

\clearpage
\section{Additional Experimental Results with UPS} \label{sec:add_exp_ups}

\begin{figure*}[!ht]
    \footnotesize
    
    \centering
    \begin{minipage}[ht]{0.49\linewidth}
    \center{\includegraphics[width=1\linewidth]{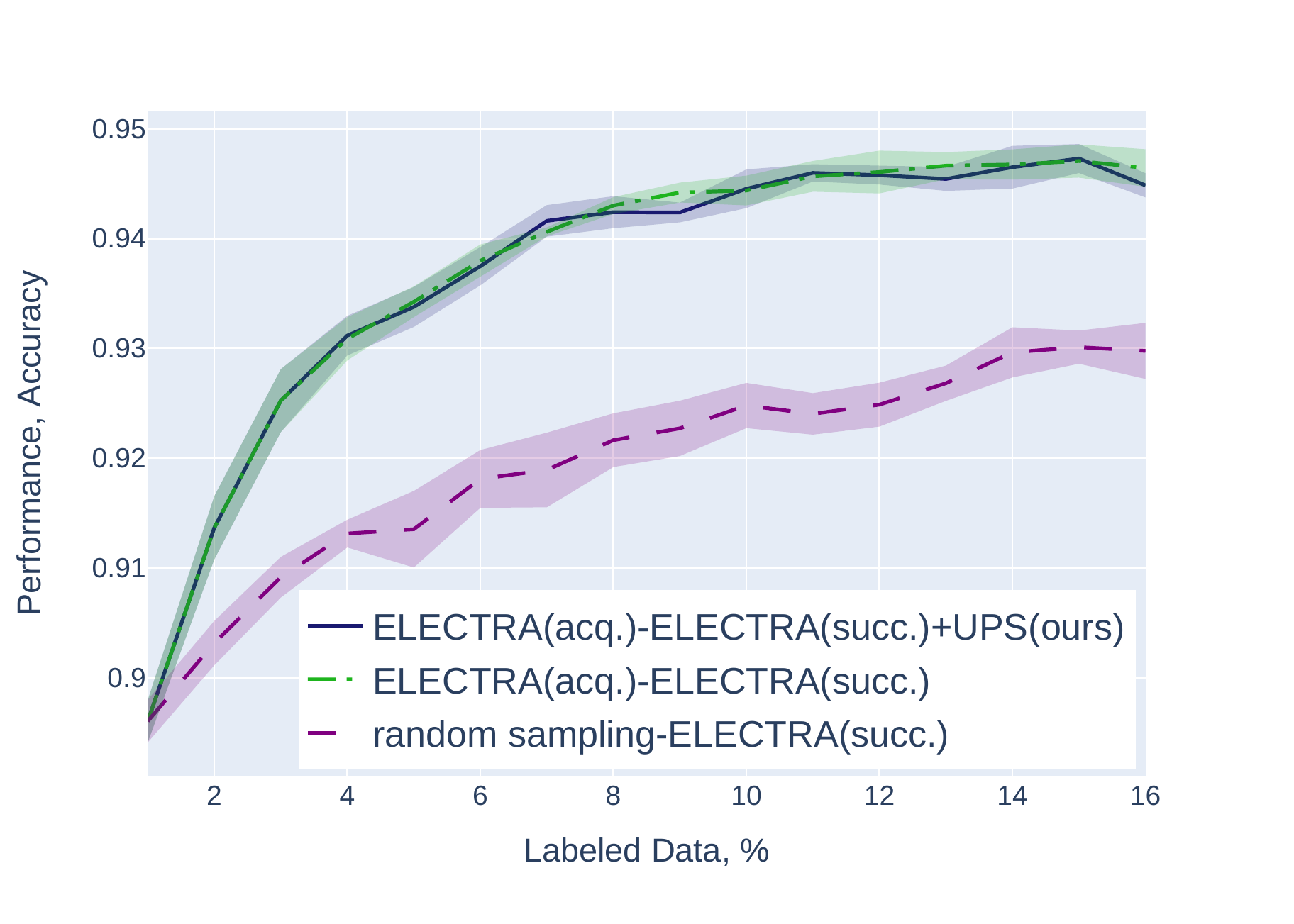} a) AG News dataset. }
    \end{minipage}
    \hspace{0.1cm}
    \begin{minipage}[ht]{0.49\linewidth}
    \center{\includegraphics[width=1\linewidth]{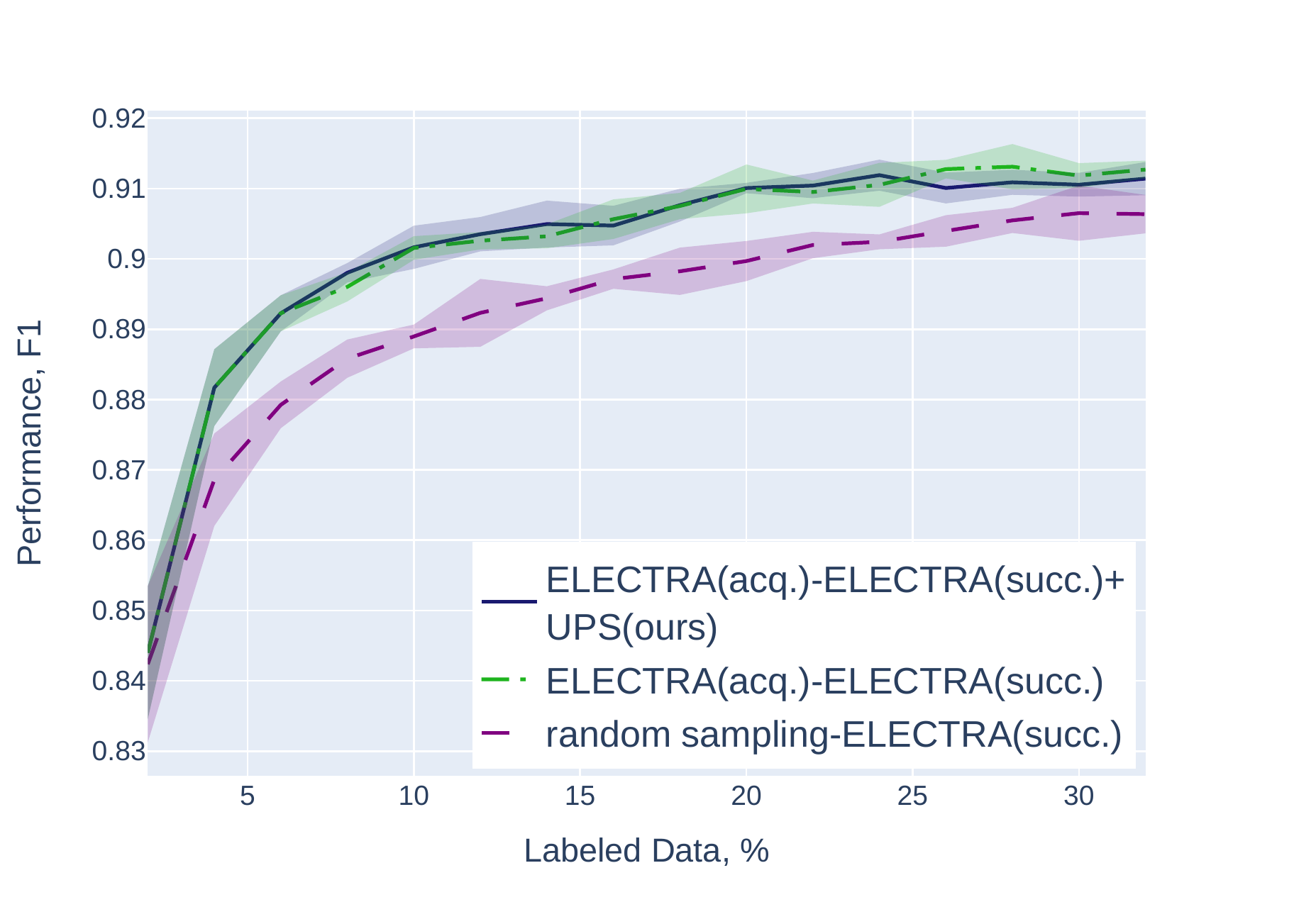} b) CoNLL-2003 dataset.}
    \end{minipage}
    
    \caption{The performance of UPS compared with the standard approach to AL on AG News and CoNLL-2003 datasets with ELECTRA as a successor model ($\gamma = 0.1$, $T=0.01$). }
    \label{fig:ups_ag_and_conll}
\end{figure*}

    
    

    
    

\begin{figure}[!ht]
	\center{\includegraphics[width=0.5\linewidth]{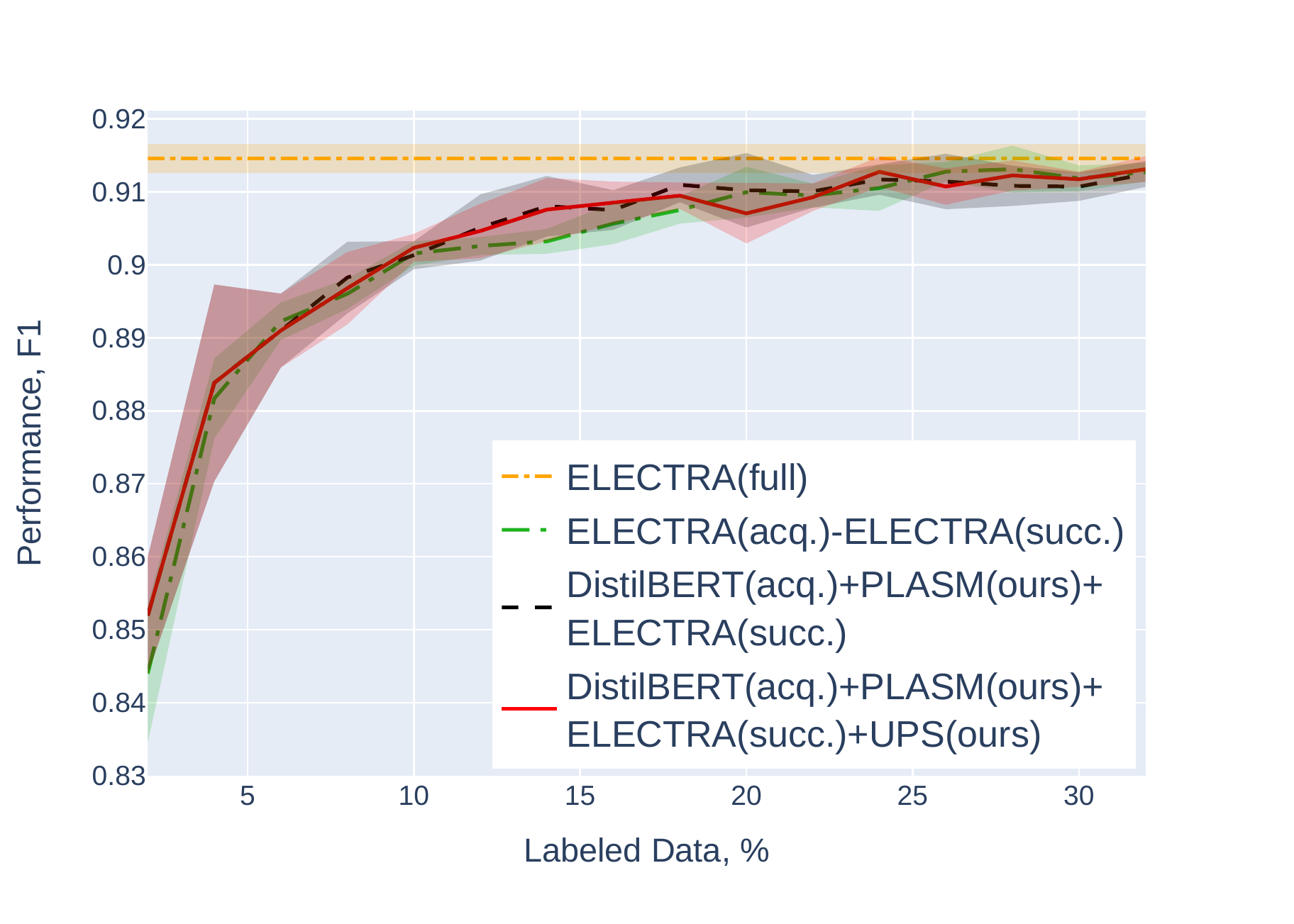}}
	\caption{The performance of UPS in conjunction with PLASM (BERT is a pseudo-labeling model) on CoNLL-2003 compared with baselines ($\gamma = 0.1$, $T=0.01$).}
	\label{fig:ups_conll}
	
\end{figure}

\begin{figure}[!ht]
	\center{\includegraphics[width=0.5\linewidth]{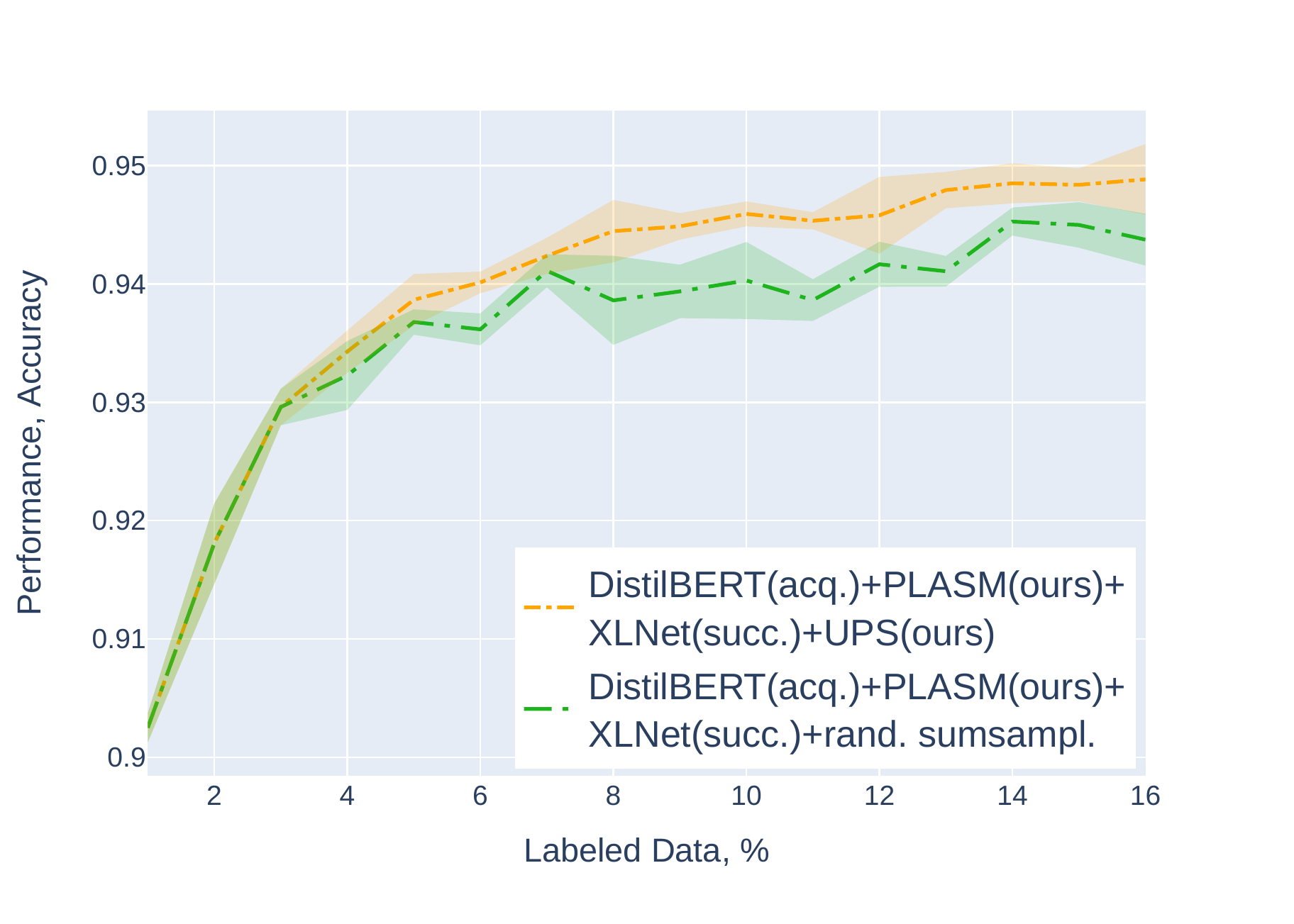}}
	\caption{The comparison of UPS with a random-subsampling baseline on the AG News dataset ($\gamma = 0.1$, $T=0.01$). A pseudo-labeling model in PLASM is BERT.}
	\label{fig:ups_random}
	
\end{figure}

\begin{figure}[!ht]
	\center{\includegraphics[width=0.5\linewidth]{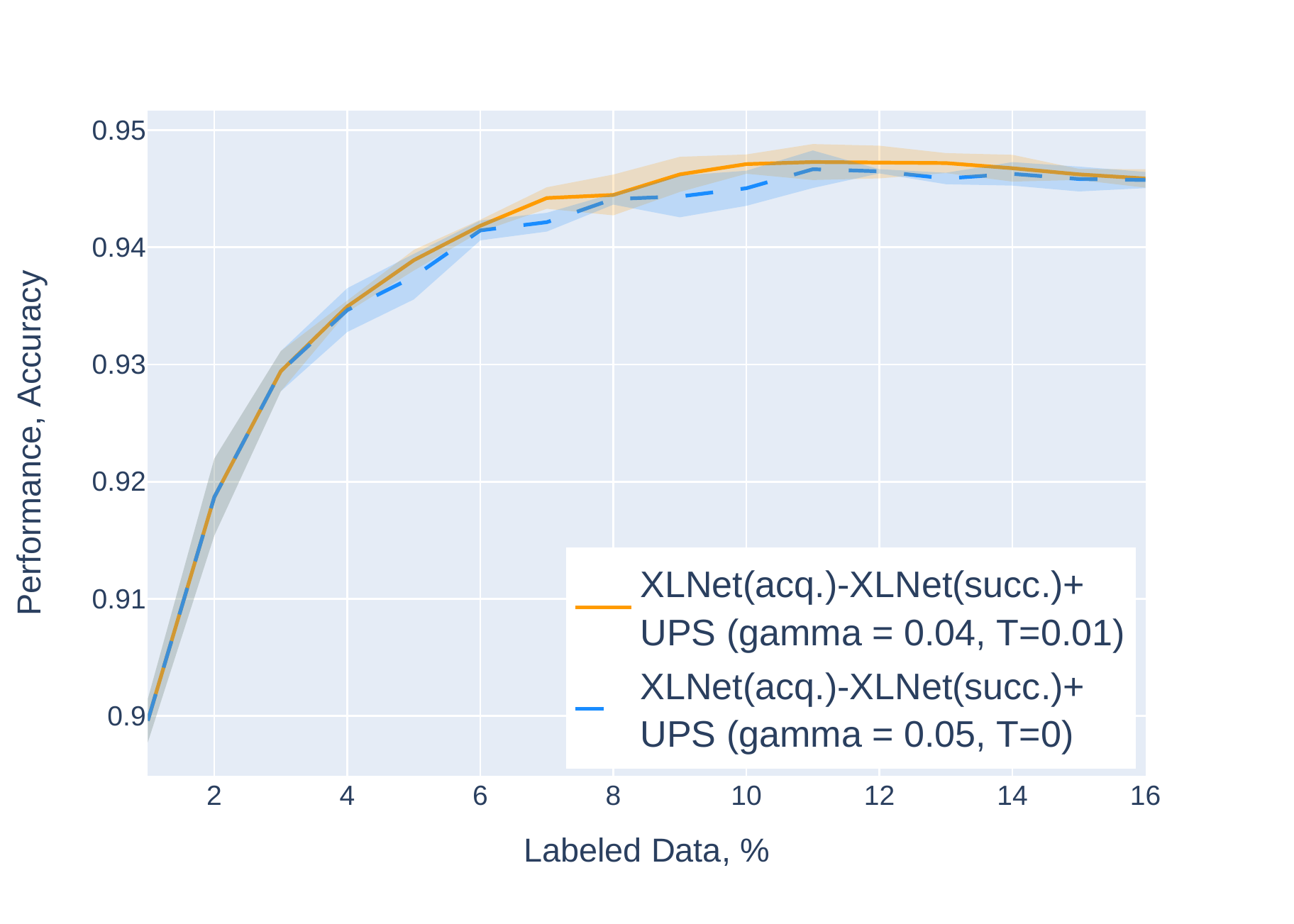}}
	\caption{Ablation study for the parameter $T$ in the UPS algorithm. We see that when sampling a total of 5\% of the dataset to select for the query, using a non-zero value for the parameter $T$ gives an increase in performance compared to a case when only 5\% most uncertain samples are considered for query (i.e. $T$ = 0). }
	\label{fig:ups_T_0}
	
\end{figure}

\begin{table*}[!ht]
\footnotesize
\centering

\scalebox{1.0}{\begin{tabular}{cllllll}
\hline
&{} &           ELECTRA &              BERT &        DistilBERT & \begin{tabular}[c]{@{}l@{}}\textbf{ELECTRA}\\ \textbf{with UPS (ours)}\end{tabular} & \begin{tabular}[c]{@{}l@{}}\textbf{DistilBERT}\\ \textbf{with UPS (ours)}\end{tabular} \\

\hline
\multirow{3}{*}{\rotatebox{90}{Iter. 2}} & Train             &     $44.8\pm 0.3$ &     $50.9\pm 1.6$ &     $29.1\pm 0.3$ &           $43.3\pm 0.8$ &              $26.4\pm 2.5$ \\
& Inference         &     $25.9\pm 0.3$ &     $25.9\pm 0.3$ &     $19.6\pm 0.3$ &           $25.7\pm 0.4$ &              $19.9\pm 0.9$ \\
& Overall           &     $70.6\pm 0.6$ &     $76.8\pm 1.7$ &     $48.7\pm 0.5$ &           $69.0\pm 1.0$ &              $46.3\pm 3.1$ \\
\hline
\multirow{3}{*}{\rotatebox{90}{Iter. 6}} & Train             &     $74.9\pm 1.6$ &     $81.4\pm 1.4$ &     $49.7\pm 1.3$ &           $66.9\pm 1.6$ &              $44.2\pm 4.0$ \\
& Inference         &     $23.8\pm 0.0$ &     $23.4\pm 0.3$ &     $17.9\pm 0.0$ &            \textbf{3.2}$\pm $\textbf{0.2} &               \textbf{2.3}$\pm $\textbf{0.2} \\
& Overall           &     $98.6\pm 1.5$ &    $104.8\pm 1.1$ &     $67.5\pm 1.4$ &           \textbf{70.1}$\pm $\textbf{1.6} &              \textbf{46.5}$\pm $\textbf{4.2} \\
\hline
\multirow{3}{*}{\rotatebox{90}{Iter. 10}} & Train             &     $95.6\pm 1.1$ &    $105.7\pm 1.5$ &     $63.6\pm 2.0$ &           $88.4\pm 1.2$ &              $57.1\pm 5.5$ \\
& Inference         &     $21.3\pm 0.1$ &     $21.4\pm 0.2$ &     $15.9\pm 0.5$ &            \textbf{2.6}$\pm $\textbf{0.2} &               \textbf{2.3}$\pm $\textbf{0.1} \\
& Overall           &    $116.9\pm 1.2$ &    $127.1\pm 1.5$ &     $79.5\pm 2.4$ &           \textbf{91.0}$\pm $\textbf{1.3} &              \textbf{59.4}$\pm $\textbf{5.6} \\
\hline
\multirow{3}{*}{\rotatebox{90}{Iter. 15}} & Train             &    $122.2\pm 1.2$ &    $133.4\pm 3.1$ &     $79.0\pm 1.3$ &          $129.9\pm 3.2$ &              $74.6\pm 6.4$ \\
& Inference         &     $18.9\pm 0.2$ &     $18.6\pm 0.1$ &     $14.0\pm 0.2$ &            \textbf{2.0}$\pm $\textbf{0.1} &               \textbf{1.4}$\pm $\textbf{0.1} \\
& Overall           &    $141.1\pm 1.0$ &    $151.9\pm 3.2$ &     $92.9\pm 1.2$ &          \textbf{131.9}$\pm $\textbf{3.1} &              \textbf{76.0}$\pm $\textbf{6.5} \\
\hline
                      \multicolumn{2}{l}{Overall train}     &  $1266.6\pm 16.9$ &  $1387.1\pm 26.3$ &   $838.6\pm 19.2$ &        $1195.0\pm 25.0$ &            $748.3\pm 70.4$ \\
\multicolumn{2}{l}{Overall inference} &    $339.1\pm 3.5$ &    $335.5\pm 4.7$ &    $252.9\pm 3.9$ &          \textbf{128.9}$\pm $\textbf{5.6} &              \textbf{97.5}$\pm $\textbf{5.1} \\
\multicolumn{2}{l}{Overall}           &  $1605.7\pm 18.8$ &  $1722.6\pm 24.1$ &  $1091.4\pm 18.4$ &        \textbf{1323.9}$\pm $\textbf{28.5} &            \textbf{845.8}$\pm $\textbf{75.1} \\
 \hline
\end{tabular}}
\caption{Duration of training and inference steps of AL iterations in seconds on CoNLL-2003. We highlight with the bold font the values affected by UPS. Hardware configuration: 2 Intel Xeon Platinum 8168, 2.7 GHz, 24 cores CPU; NVIDIA Tesla v100 GPU with 32 Gb of VRAM.
}
\label{tab:time_conll}

\end{table*}

\clearpage

    
    

\end{document}